\documentclass[runningheads]{llncs}
\usepackage{eccv}
\usepackage{eccvabbrv}
\usepackage{graphicx}
\usepackage{booktabs} 
\usepackage[pagebackref,breaklinks,colorlinks,citecolor=eccvblue]{hyperref}
\usepackage{adjustbox}
\usepackage{bm}
\usepackage{multirow}
\usepackage{bbding}

\begin{document}

\title{Spread Your Wings: A Radial Strip Transformer for Image Deblurring} 

\titlerunning{A Radial Strip Transformer for Image Deblurring}

\author{Duosheng Chen\inst{1,2}\and
Shihao Zhou\inst{1,2} \and
Jinshan Pan\inst{3} \and
Jinglei Shi\inst{1} \and \\
Lishen Qu\inst{1}\and
Jufeng Yang\inst{1,2} 
}
\authorrunning{Chen et al.}
\institute{VCIP \& TMCC \& DISSec, College of Computer Science, Nankai University \and
Nankai International Advanced Research Institute (SHENZHEN· FUTIAN) \and
School of Computer Science and Engineering, Nanjing University of Science and Technology}

\maketitle
\begin{abstract}
Exploring motion information is important for the motion deblurring task. 
Recently window-based transformer approaches have achieved decent performance in image deblurring. Note that the motion causing blurry results is usually composed of translation and rotation movements and the window-shift operation in the Cartesian coordinate system by the window-based transformer approaches only directly explores translation motion in orthogonal directions. Thus, these methods have the limitation of modeling the rotation part.
To alleviate this problem, we introduce the polar coordinate-based transformer, which has the angles and distance to explore rotation motion and translation information together.
In this paper, we propose a \textbf{R}adial \textbf{S}trip \textbf{T}ransformer (RST), which is a transformer-based architecture that restores the blur images in a polar coordinate system instead of a Cartesian one.
RST contains a dynamic radial embedding module (DRE) to extract the shallow feature by a radial deformable convolution.
We design a polar mask layer to generate the offsets for the deformable convolution, which can reshape the convolution kernel along the radius to better capture the rotation motion information.
Furthermore, we proposed a radial strip attention solver (RSAS) as deep feature extraction, where the relationship between windows is organized by azimuth and radius.
This attention module contains radial strip windows to reweight image features in the polar coordinate, which preserves more useful information in rotation and translation motion together for better recovering the sharp images.
Experimental results on six synthetic and real-world datasets prove that our method performs better than the SOTA methods for the image deblurring task.
\textit{} {The code are available in the \href{https://github.com/Calvin11311/RST}{https://github.com/Calvin11311/RST}.}
\end{abstract}

\section{Introduction}
\label{sec:intro}
Image deblurring aims to recover a sharp image from the blurred one. This is a challenging problem as both blur and latent clear images are unknown.  
This problem has attracted many researchers from academia and industry due to the rapid development of kinds of smartphones.
Significant progress has been made due to kinds of deep convolution neural networks (CNNs)~\cite{zamir2021multi,nah2017deep,fang2023self}.
The key success of these methods is because of the capability to model the local blur pattern by a convolution kernel and the diversity design of network architecture (\eg, the multi-scale~\cite{zamir2021multi,zhang2019deep}, multi-stage~\cite{nah2017deep,tao2018scale} generative adversarial learning~\cite{Kupyn_2018_CVPR,kupyn2019deblurgan}, and latent space kernel estimation~\cite{fang2023self}).
Since the limited receptive field and the spatially invariant of the convolution kernels~\cite{zamir2022restormer} can not fully explore non-local contextual information that is vital for image deblurring, a deeper and larger network may relieve this problem. However, simply increasing the depth or capacity of deep networks does not always lead to better performance~\cite{zhang_deblur_cvpr18,pan2020physics}.

\begin{figure*}[t]
\scriptsize
\centering
\begin{adjustbox}{valign=t}
\begin{tabular}{c}
\includegraphics[width=0.9\textwidth]{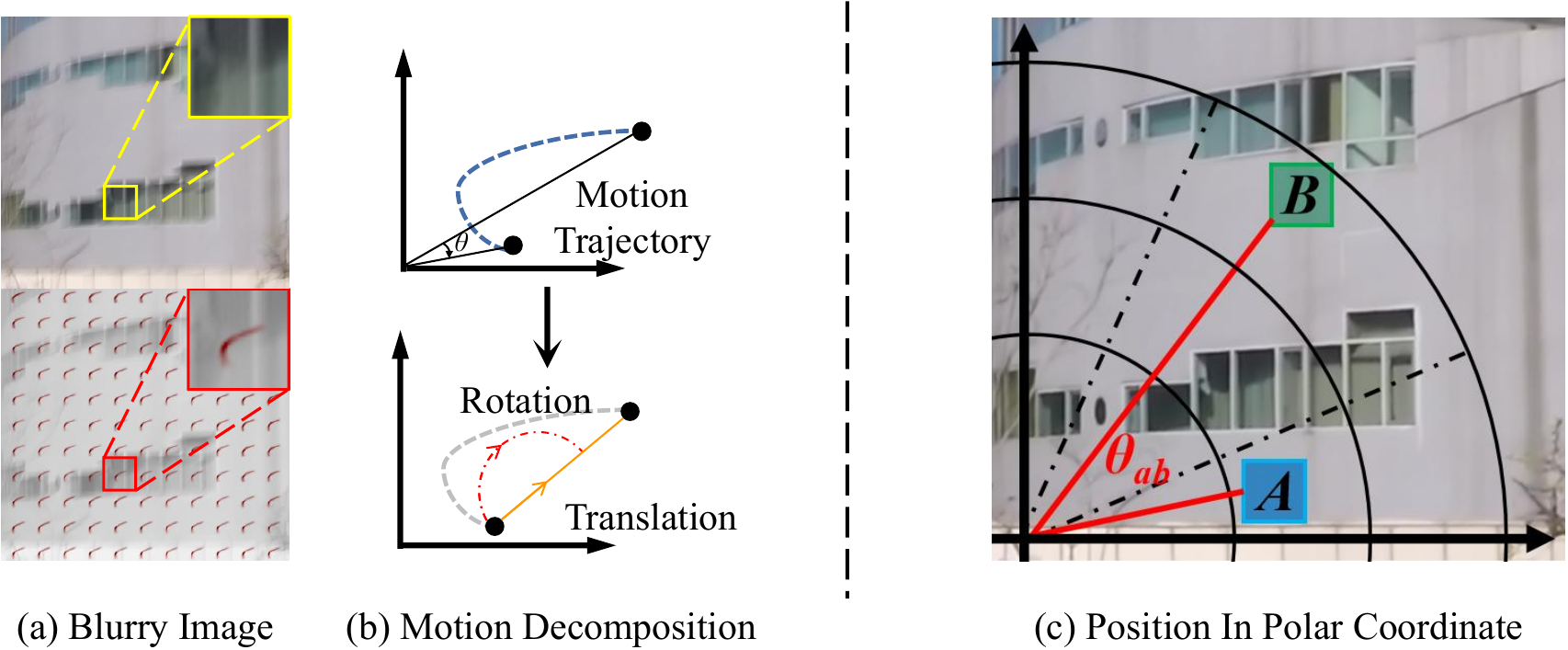} 
\end{tabular}
\end{adjustbox}
\vspace{-2mm}
\caption{
(a) Illustration of the motion field from the kernel estimation method~\cite{carbajal2023blind}, the red curves describe the motion state which causes the blurry images.
(b) The demonstration of the motion trajectory causing blurry results can be composed of translation and rotation parts, as the blue curve consists of the red rotation motion and the yellow translation motion.
(c) Illustrate the token relationship modeled under the polar coordinate system, where the $\theta_{ab}$ can represent the angular relative position between tokens (pixels) A and B.
}
\label{pic:fig1}
\end{figure*}
To overcome this problem, several approaches~\cite{zamir2022restormer,eccv2022_Stripformer,kong2023efficient} explore the vision Transformer~\cite{dosovitskiy2020image} to solve image deblurring and achieve decent performance on kinds of benchmark datasets. 
However, the scaled dot-product attention in the Transformer is time-consuming.
Some methods adopt the rectangle window~\cite{chen2022cross} or the strip windows~\cite{eccv2022_Stripformer} to compute self-attention on lower computation complexity.
Since the restricted spatial extent of self-attention results in limitations for capturing long-range pixel relationships, the window shift operation is necessary to expand the receptive field.
The shift operation under the Cartesian coordinate system captures the relationship between windows in orthogonal directions.
However, in Figure~\ref{pic:fig1} (a), we can clearly find that the blur kernel in real-world blurry images is an irregular curve indicating that the motion causing blurry images is complex.
Furthermore, the motion can usually be decomposed into rotation and translation motion.
Deng~\etal~\cite{qiu2019world} formulate the decomposition as $RT(o,\theta,t,x)=R(\theta)(x-o)+o+t$, where t and $\theta$ represent the translation vector and the rotation angle, respectively, $o$ and $x$ are the coordinates in the blurry image and the $R(\theta)$ is a geometric rotation matrix.
Based on the above theory, as illustrated in Figure~\ref{pic:fig1} (b), the trajectory of motion can consist of the rotation (the red curve) and the translation (the yellow line).
We note that attention operations capturing both translation and rotation motion information together can enhance the ability to restore clearer images.

However, shift operation under the Cartesian coordinate system (\eg, Stripformer~\cite{eccv2022_Stripformer}) only considers the cross-window connection in orthogonal directions, which explicitly models translation motion, while implicitly modeling rotation motion, leading to the limited capability to model the rotation information for restoring sharp images.
In the polar coordinate system, we can use angles and distance to demonstrate the relative position, where the translation and the rotation can be considered together.
So it is more suitable to develop an effective and efficient polar system-based transformer architecture to explore the motion information of blurry images better.
In this paper, we propose a \textbf{R}adial-\textbf{S}trip \textbf{T}ransformer (RST), exploiting two polar coordinate system-based modules: the dynamic radial embedding (DRE) and radial strip attention solver (RSAS), to capture the rotation and the translation features of motion efficiently.
The DRE utilizes a polar mask to construct the shallow feature extractor, which is combined with the deformable convolution to capture the motion information of blur on the polar way.
In detail, the rotation information can be captured by angles and the translation part is modeled by distance.

Since features extracted from DRE are not fully utilized by attention operations with the Cartesian coordinate system,
directly using window-based transformer architecture (\eg, Swin Transformer~\cite{liu2021swin}) will not result in a good performance of image deblurring.
Inspired by the DarSwin~\cite{athwale2023darswin}, which proposes a polar transformer under a wide-angle lens, 
we design an efficient radial strip attention solver (RSAS), which contains strip windows along the radius.
Sampling operations are needed in sector window partitioning but can result in the loss of details, which is important for image deblurring.
So we cancel the sampling operation and set the strip window, which is beneficial for preserving the useful detailed information for clear image restoration.
Combined with the two designs above, we proposed our RST with an asymmetric encoder-decoder architecture, which only uses the RSAS in the decoder part.
It is more efficient to solve the deblurring task because the shallow layers in attention may dilute the features of blur pattern~\cite{kong2023efficient}.
The experimental results support that the method performs a favorable deblurring result against SOTA methods.
The title ``Spread your wings" gives us the imagination demonstrating a process of exploring more directions for capturing the motion information more fully under the polar coordinates.

Our main contributions in this paper can be summarized as follows:
\vspace{-2mm}
\begin{itemize}
\item[$\bullet$]
We propose an efficient polar coordinate system-based embedding module DRE for extracting rotation and translation motion information of shallow features.
It transforms the deformable convolution layer in each radial direction and preserves the information in rotation and translation. 
%
\item[$\bullet$]
We design a new polar attention module RSAS radial strip window without dropping the useful detail information for clear image restoration.
\item[$\bullet$]
We evaluate our RST on six synthesis and real-world datasets, showing that our method performs favorably against state-of-the-art methods.
\end{itemize}

\section{Related Work}
\label{sec:Related_Work}

\subsection{Deep CNN-based approaches}
Recently, the development of deep CNN methods~\cite{chen2022simple,cvpr2021cho,gao2019dynamic} has achieved great performance in the image deblurring task.
Nah~\etal~\cite{nah2017deep} design a muti-scale network to estimate the blur pattern from a blurry image directly.
To make the multi-scale architecture extract the blur pattern more efficiently, Gao~\etal~\cite{gao2019dynamic} propose a parameter selective sharing method.
\cite{sun2015learning} utilized the CNNs to predict the probabilistic distribution of the blurring kernel at the patch level. 
Moreover, the multi-patch architecture is another type to improve the performance of deblurring.
Zhang \etal~\cite{zhang2019deep} perform a hierarchical deblurring model based on the multi-patch strategy, which estimates the feature of the blur pattern stage by stage.
The cross-stage strategy is utilized by Zamir~\etal~\cite{zamir2021multi}, which fuses the different stage features to better restore the sharp image.

The above methods only directly use the fixed convolution kernel to estimate the blur pattern, which ignores the important movement prior of the blurry images. 
Some methods consider the direction of degradation, and 
WDNet~\cite{liu2020wavelet} designs a dual-branch network for demoir{\'e}ing, which contains a DPM block to capture the texture from eight directions.
DSC~\cite{hu2019direction} utilizes the attention mechanism in a spatial recurrent neural network for locating shadows from each direction.
Fang~\etal~\cite{fang2022uncertainty} try to estimate the non-uniform blur kernel in latent space and use it to reconstruct the sharp image.
Due to the convolution operation being a spatial invariant and limited receptive field~\cite{zamir2022restormer}, CNN methods do not efficiently capture the global information for restoring blurry images.
\subsection{Vision Transformer-based approaches}
Transformer architecture achieves a significant performance of modeling global contexts with self-attention in high-level tasks (\eg, moving object segmentation~\cite{pan2019joint}, object detection~\cite{Kupyn_2018_CVPR}, and text recognition~\cite{lee2019blind}), and extend to low-level tasks(image deblurring~\cite{kong2023efficient,zamir2022restormer}, super-resolution~\cite{liang2021swinir}, and image denoising~\cite{chen2021pre,wang2022uformer}).
But, self-attention brings a problem of huge computational cost.
To solve it, SwinIR~\cite{liang2021swinir} divides the image by 8$\times$8 windows and executes an attention operation in each window. 
Restormer~\cite{zamir2022restormer} designs across channel dimensions model to reduce the cost.
The spatial information is important for the image restoration task, which is not fully developed.
CAT~\cite{chen2022cross} aggregates the features across rectangle windows with axial-shift operation, which can achieve a large receptive field and boost the interactions between windows with linear complexity.
Stripformer~\cite{eccv2022_Stripformer} proposed an intra-strip and inter-strip attention mechanism, which decomposes motion information into horizontal and vertical directions for estimating the blur pattern.
However, the above methods designed based on the Cartesian coordinate system can only explore the motion information in orthogonal directions.
Since translation and rotation are both important parts of object movement, the Cartesian coordinate has a limited ability to model the rotation.
Different from vanilla transformers calculate the relationship between pixels on the Cartesian system, we propose an efficient polar coordinate transformer model to explore the spacial information of movement with translation and rotation.

\begin{figure*}[!t]
\scriptsize
\centering
\begin{adjustbox}{valign=t}
\begin{tabular}{c}
\includegraphics[width=1\textwidth]{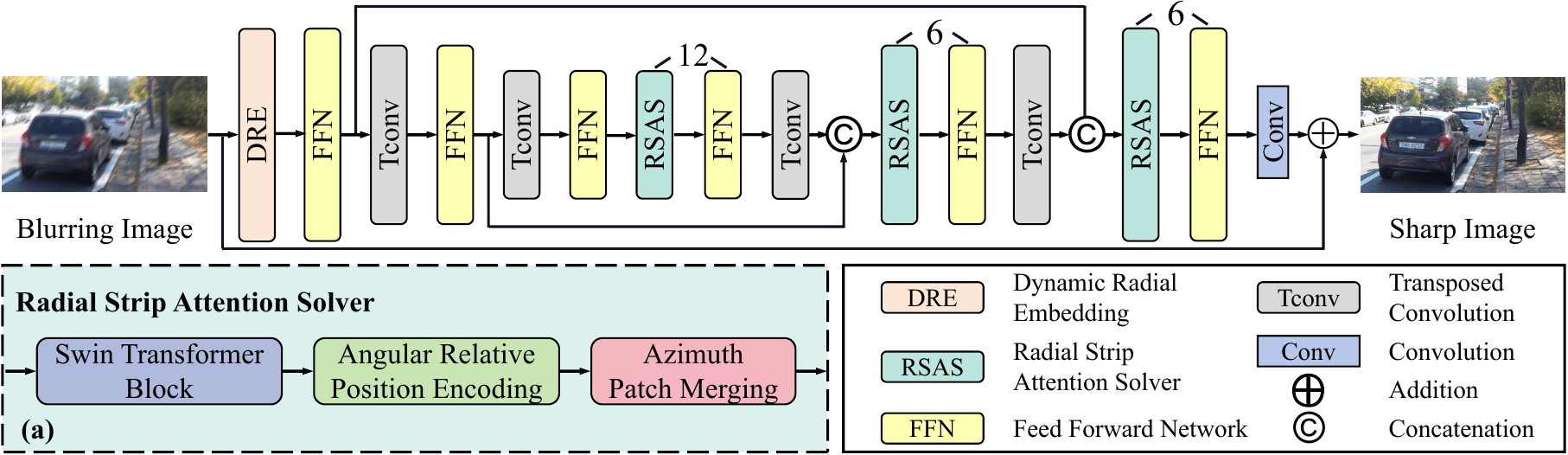} \hspace{0mm} 
\\
\end{tabular}
\end{adjustbox}
\\
\begin{adjustbox}{valign=t}
\end{adjustbox}
\caption{
Overview of RST. 
RST consists of an asymmetric encoder-decoder architecture and the encoder module only has FFN.
The proposed dynamic radial embedding (DRE) module extracts shallow features under the polar coordinate system. 
(a) Illustration of the radial strip attention solver (RSAS) consisting of window-based attention and angular relative positional encoding. 
}
\label{pic:pipeline}
\end{figure*}

\section{Proposed Method}
To restore high-quality blurry images effectively and efficiently, we propose an efficient radial strip transformer method to model the motion from rotation and translation.
To this end, we propose a dynamic radial embedding module as the shallow feature extraction, which consists of a radial mask and a deformable convolution~\cite{dai2017deformable} layer.
Furthermore, the attention module is consistently designed on the polar coordinate with the radial strip window and relative angular position encoding.
In our method, the Feed-Forward Network (FFN) is frequency domain-based, owing to the FFN design of FFTformer~\cite{kong2023efficient}, aiming to preserve more useful information for restoring sharp images. 
Then, we formulate the proposed modules into 
an asymmetric encoder-decoder architecture, which cancels the attention block in the encoder, following the~\cite{kong2023efficient}.

The overall pipeline of our proposed RST is presented in Figure~\ref{pic:pipeline}. 
We demonstrate two proposed blocks: the dynamic radial embedding later described in Section~\ref{method:polarembedding}, and the radial strip attention solver illustrated in Section~\ref{method:PA}.

\begin{figure*}[!t]
\scriptsize
\centering
\begin{adjustbox}{valign=t}
\begin{tabular}{ccc}
\includegraphics[width=0.25\textwidth]{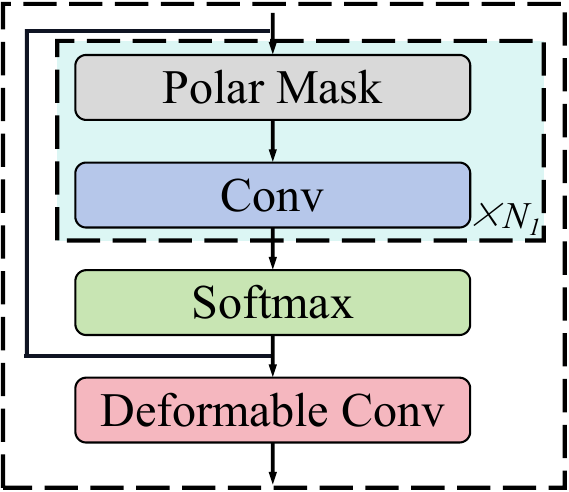} &\hspace{0.8cm}
\includegraphics[width=0.5\textwidth]{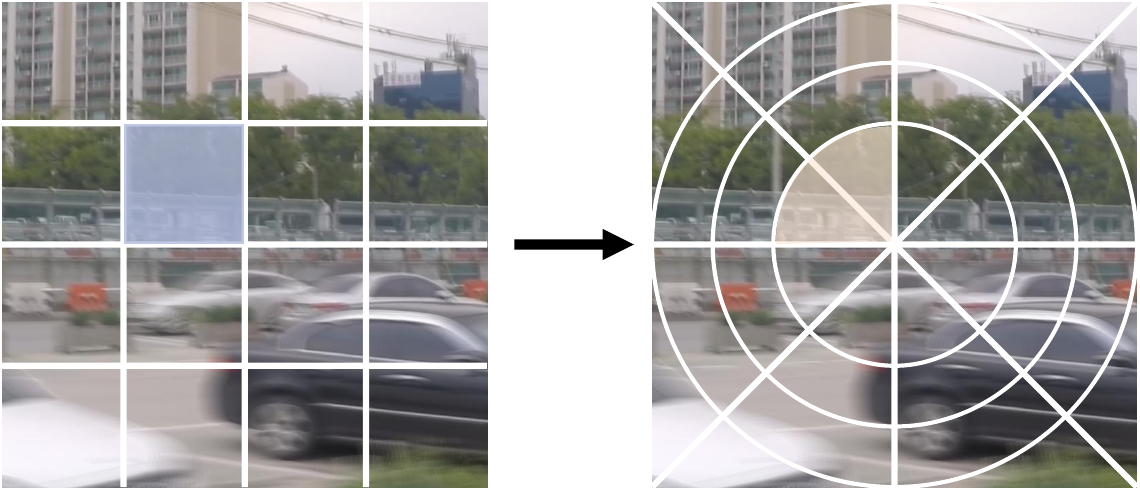} &\hspace{0cm}
\\
(a) DRE architecture &\hspace{0cm}
(b) Embedding Transformation &\hspace{0cm}
\\
\end{tabular}
\end{adjustbox}
\caption{
(a) Illustration of dynamic radial embedding. 
The polar mask and the convolution layer compose the offset for the next deformable convolution layer.
${N}_{1}$ represents the number of sectors we split, we finally add a softmax layer to normalize the offsets of each sector.
(b) Illustration of the difference between Swin~\cite{liu2021swin} transformer embedding module (left) and our radial embedding (right). 
Instead of the Cartesian system and CNNs extracting the shallow feature of the input image, we use a polar coordinate and radial deformable convolution to capture the features.
In our case, the radial embedding strategy makes convolution operation along the azimuth.
We use the white borders to denote the divisions of the windows and the blur and the yellow region shows the areas being captured along the horizontal and azimuth. 
}
\label{pic:fig3}
\vspace{-4mm}
\end{figure*}

\subsection{Overall Pipeline}
Given a blurry image $\bm{X}$ with a resolution of $ H\times W$.
RST first adopts a deformable convolution layer with a radial mask to extract the shallow feature $\bm{F}_{0} \in \mathbb{R}^{H\times W\times C}$, where $H,~W,~C$ denote the height, width, and number of channels, respectively.
Then the shallow feature $\bm{F}_{0}$ is embedded in the L-level encoder-decoder networks where each level has a skip connection between the encoder and decoder.
Following the pioneered work~\cite{kong2023efficient}, 
we directly use the frequency domain-based feed-forward network to refine the features, which has the benefit of 
capturing useful context information for deblurring.
The process can be formulated as:
\begin{equation}
\bm{F}_{e}=\textbf{FFN}(\mathcal{F}^{-1}(\bm W \mathcal{F}(\bm{F}_{0}))),
\label{eq:fft_ffn}
\end{equation}
$\mathcal{F}(\cdot)$ denotes the fast Fourier transform and $\mathcal{F}^{-1}(\cdot)$ denotes the inverse one. 
$\bm W$ represents a learnable quantization matrix.
$\bm{F}_{e} \in \mathbb{R}^{H\times W\times C}$ represents the output feature from \textbf{FFN} of encoder parts.
We use the transpose convolution for downsampling and upsampling and add the skip connection between the encoder and decoder on each level.
In the decoder, we design radial strip attention (as shown in Figure~\ref{pic:pipeline}) to extract the blur pattern from the rotation and translation movement.
To be specific, the feature $\bm{F}_{a} \in \mathbb{R}^{\frac{H}{M}\times \frac{W}{M}\times C}$ ($M$ represents the patch size) product by window partition layer.
After that, the windows are assembled by an azimuth patch merging module with the angular relative positional encoding.
The transformer block outputs the final restored feature $\bm{F}_{r} \in \mathbb{R}^{H\times W\times C}$, and using a $3 \times 3$ convolution layer to generate the residual part $\bm{R} \in \mathbb{R}^{3\times H\times W}$.
Finally, the restored image is generated by $\bm{X}^{'} = \bm{X} + \bm{R}$.
\subsection{Dynamic Radial Embedding}
\label{method:polarembedding}
The first step of our proposed transformer architecture is embedding the patches in the polar coordinate system.
In previous works~\cite{eccv2022_Stripformer,kong2023efficient,zamir2022restormer}, the embedding layer often consists of a $3 \times 3$ convolution to extract the shallow features of input images. 
However, the fixed kernel and square-shaped receptive field limit the capability of obtaining the real motion information of the blurry image, which ignores the rotation.
To alleviate this problem, as opposed to the Swin transformer which splits the patches based on a Cartesian coordinate, we designed a new polar coordinate-based embedding module DRE, which consists of four layers: polar mask, convolution, softmax, and deformable convolution.

\begin{figure*}[t]
\scriptsize
\centering
\begin{adjustbox}{valign=t}
\begin{tabular}{ccc}
\includegraphics[width=0.23\textwidth]{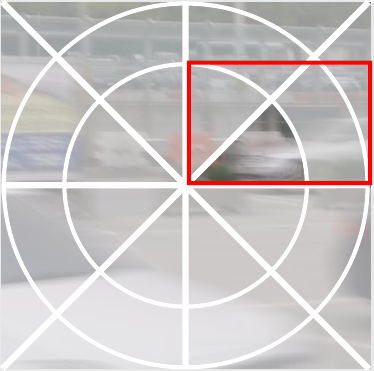} &\hspace{0cm}
\includegraphics[width=0.35\textwidth]{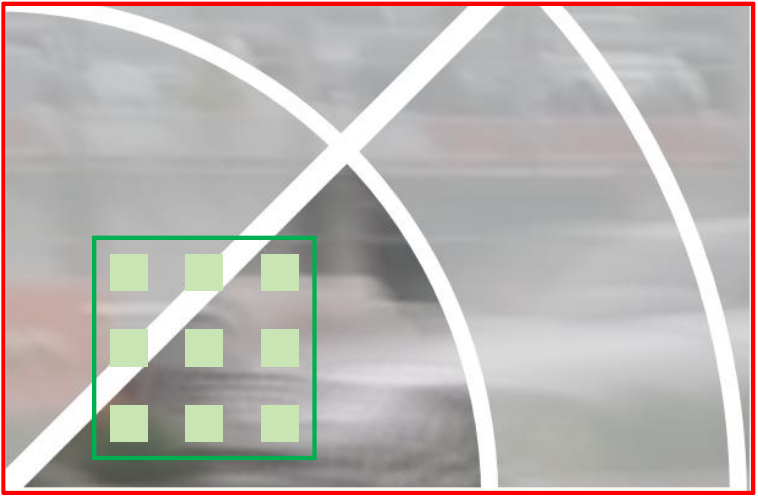} &\hspace{0cm}
\includegraphics[width=0.35\textwidth]{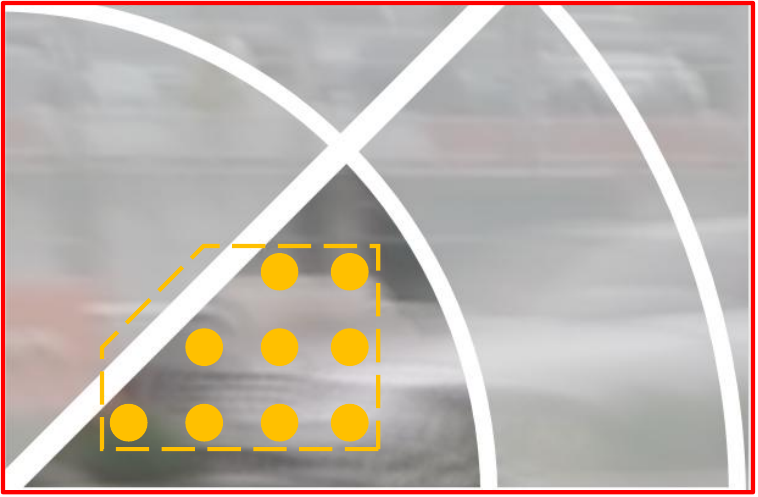}
\vspace{1mm}
\\
(a) Blurry input &\hspace{0cm}
(b) Normal grid kernel &\hspace{0cm}
(c) Polar deformable kernel 
\vspace{1mm}
\\
\end{tabular}
\end{adjustbox}
\caption{
(a) Illustration of the polar embedding. 
In the polar embedding patch, since the shape of patches is a sectorial region, the boundary consists of curves and slanting lines rather than a horizontal or vertical boundary.
(b) Illustration of convolution layer.
We demonstrate the convolution kernel size by the green square dots, it has the fixed shape($e.g. 3 \times 3$) of the convolution kernel.
Since the slanting edge of the patch, the upper-left square dots of the normal grid kernel are out of range, resulting in the convolution of these pixels in this portion becoming ineffective.
(c) Illustration of deformable convolution layer.
We use the yellow round dots to represent the receptive field of the deformable convolution.
Based on the offsets, the area of the receptive field is reshaped with the structured information of the patches, which is sectorial, not squared.
Since the adaptability of the deformable convolution to flexible shapes, it can capture shallow features better than normal convolution within the masked region.
}

\label{pic:fig4}
\vspace{-4mm}
\end{figure*}

As shown in Figure~\ref{pic:fig3} (a), 
the DRE module consists of a deformable convolution, we generate the offsets of the deformable convolution by the polar mask layer and convolution layer.
Given the input image is $\bm{X} \in \mathbb{R}^{H\times W\times 3}$, where the H, W, and 3 represent the height, weight, and the number of channels.
Firstly, we generate a distribution matrix $\bm{D}_{c}\in \mathbb{R}^{H\times W}$ 
based on the size of the input image, which demonstrates the position of each pixel in the Cartesian coordinate system.
Then, we apply the conversion rule between the Cartesian coordinate and polar coordinate, which can be formulated as:
\begin{equation}
\bm{D}_{\theta}={arctan}(\frac{\bm D_{y}}{\bm D_{x}}),
\label{eq:transco}
\end{equation}
$\bm {D}_{\theta} \in \mathbb{R}^{H\times W}$ denotes the distribution matrix of polar coordinate and ${arctan}$ denotes the arctangent function. 
$\bm D_{x}, \bm D_{y} \in \mathbb{R}^{H\times W}$ represents position matrices of the x-axis and the y-axis based on the Cartesian coordinate.
Then, we use a polar mask to generate the offsets in each sector part by convolution layer.
Given the ${N}_{1}$ as the number of azimuths in the polar coordinate system, we divided the full image into ${N}_{1}$ parts and generated the corresponding mask matrix.
We combine the mask matrix with the convolution layer and capture the offsets from the directions of each azimuth.
\begin{equation}
\Delta{p}_{i}= Conv_{3\times3}(\bm{X} \cdot \bm{D}_{\theta \_{i}})\quad i = \{1, 2, ..., N_{1}\}
\label{eq:offset}
\end{equation}
where the $Conv_{3\times3}(\cdot)$ denotes the $3 \times 3$ convolution layer, and the mask matrix of sector $i$ is represented as $\bm{D}_{\theta \_{i}})$.
$\Delta{p}_{i} \in \mathbb{R}^{H\times W\times G}$ denotes the offsets tensor in sector $i$, where $G$ means the number of offsets groups, equaling to $2\times kernel\_size \times kernel\_size$.
After ${N}_{1}$ times features capturing along the radial, we generate the final offsets by element-wise adding each masked convolution tensor.
The final offsets can be represented as $\Delta p \in \mathbb{R}^{H\times W\times G}$.
Although in other works~\cite{kong2023efficient} perform that CNNs~\cite{zeiler2014visualizing} can capture low-level features(\eg, texture and edges) in the shallow layers.
Unfortunately, the kernel size(\eg, $3 \times 3$) is usually fixed in the previous works, which results in the inability to capture the features by the structured information~\cite{xu2019learning}.
Moreover, the above different types of convolution still have the problem of fixed receptive field shape, which cannot explore the full context of the object in the original scene.
Next, in order to consider more information about the motion along the radial and angles, we adopt a deformable convolution~\cite{dai2017deformable} to calculate the shallow features, as shown in Figure\ref{pic:fig4}.
So we design a deformable convolution layer with radial offsets as the feature embedding layer.
The offsets tensor $\Delta {p}$ and the original image $\bm{X}$ as the input for the deformable convolution.
Finally, we generate the output feature $\bm{F}_{l} \in \mathbb{R}^{H\times W \times C}$ of DRE, the process can be formulated as:
\begin{equation}
\bm{F}_{l}=DConv(\bm{X},\Delta {p}),
\label{eq:dconv}
\end{equation}
where the $Dconv$ means the deformable convolution layer, and $\bm{F}_{l}$ denotes the output of the DRE.

\subsection{Radial Strip Attention Solver}
\label{method:PA}
As mentioned in Section~\ref{sec:intro}, vanilla transformer methods adopt the window split and the shift operation in the Cartesian coordinate system, which has the limitation of modeling the rotation motion context for restoring sharp images.
We propose the Radial Strip Attention Solver (RSAS), which consists of an angular position encoding and azimuth patch merging, as shown in \ref{pic:pipeline} (a).

{\flushleft \textbf{Window-based self attention.}}
Given the input feature is $\bm{F}_{l}$ from the embedding module, we first use a Swin transformer block, which maintains the strategy of the Swin transformer and reshapes the input feature into $\frac{HW}{M^2} \times M^2 \times C$ size, and $\frac{HW}{M^2}$ means the number of windows.
Since RST extracts the shallow features on polar coordinate by DRE, which is incompatible with the window partitioning of the Swin transformer~\cite{liu2021swin}.
The result of Figure~\ref{pic:abl_swin} also validates our assumption.
So, we use the $M_{\phi}$ and $M_{r}$ to represent the non-overlapping local window along the azimuth and radius on the polar coordinate system.
Inspired by Stripformer~\cite{eccv2022_Stripformer}, we design a new window-based self-attention containing the radial strip window shape with the size of $M_{\phi} = HW$and $M_{r} = 1$ to enhance the ability to capture the blur pattern along the radius and degrees.

{\flushleft \textbf{Angular Relative Position Encoding.}}
Swin transformer relative positional encoding contains the position bias $\bm{B}\in \mathbb{R}^{{M^{2}} \times {M^{2}}}$, $M$ denotes the number of patches in a window.
The query, key, and value matrices $Q, K, V$ are from the projections matrices $P_Q, P_K, P_V$, shared across windows.
Then the attention module computing is formulated as:
\begin{equation}
\bm{Att}(Q, K, V) = \mathrm{Softmax}\left(QK^T/\sqrt{d}+\bm {B}\right)V,
\label{eq:swin_att}
\end{equation}
where ${Q, K, V}\in \mathbb{R}^{{M^{2}} \times {d}}$ denote the queries, keys, and values, respectively; $d$ represents the dimension of query and key.

The position bias under the Cartesian system captures the relative position from orthogonal directions.
However, this design results in it only being able to directly express the translation motion, while ignoring the rotation information to capture the motion context.
To explore the rotation part of motion context for restoring the sharp image better, we propose an angular relative positional encoding to model the spatial relationship between tokens by angles.
On the angular relative positional encoding, the incident $\theta$ and the azimuth $\phi$ are utilized to capture the relative position between tokens.
Based on the two variables $\theta$ and $\phi$, we separate the position bias $\bm B$ into two parts: $\bm B_{\phi}$ and $\bm B_{\theta}$, which represent the incident-angular and azimuth-angular relative position bias respectively.
The token $i$-th on the polar coordinates can be shown as:
\begin{equation}
\begin{aligned}
\theta_{i} = \frac{\theta_{max}(i-0.5)}{N_r}, \quad \phi_{i} = \frac{2\pi(i-0.5)}{N_\phi},
\label{eq:angle}
\end{aligned}
\end{equation}
where $\theta_{i}$ and $\phi_{i}$ denote the angular of $i$-th token, and the $\theta_{max}$ represents the half field of tensor.
$N_\phi$ and $N_r$ denote the division number of angles and the length in the radial direction separately.
Then, the tokens $i$ and $j$ has the relative position $(\Delta \theta, \Delta \phi)$ can be given as:
\begin{equation}
\begin{aligned}
\Delta \theta = \theta_{i} - \theta_{j}, \quad i,j = \{1, 2, ..., N_{\theta}\}, \\
\Delta \phi = \phi_{i} - \phi_{j}, \quad i,j = \{1, 2, ..., N_{\phi}\}, 
\label{eq:delta}
\end{aligned}
\end{equation}
The $N_\theta$ represents the cut numbers in the radial direction. 
Then, the two position bias tensors $\bm B_{\theta}$ and $\bm B_{\phi}$ can be defined as:
\begin{equation}
\begin{aligned}
\bm B_{\theta} = a_{\Delta \theta}sin(\Delta \theta)+b_{\Delta \theta}cos(\Delta \theta), \\
\bm B_{\phi} = a_{\Delta \phi}sin(\Delta \phi)+b_{\Delta \phi}cos(\Delta \phi), 
\label{eq:coord_trans}
\end{aligned}
\end{equation}
where $a_{*}$ and $b_{*}$ means the parametrization of $\hat{B}_{\phi} \in \mathbb{R}^{(2{M_{\phi}-1}) \times {2}}$ and $\hat{B}_{r} \in \mathbb{R}^{(2{M_{\theta}-1}) \times {2}}$.
$M_{\phi}$ and $ M_{\theta}$ denote the number of patches in a window.
So the angular relative position encoding tensors are ${B}_{\phi}$ and ${B}_{\theta} \in \mathbb{R}^{{M_{\phi}^2} \times {M_{r}^2}}$, the whole process of attention module can be formulated as:
\begin{equation}
\bm{Att}(Q, K, V) = \mathrm{Softmax}\left(QK_T/\sqrt{d}+{B}_{\phi}+{B}_{\theta}\right)V,
\label{eq:dat_att}
\end{equation}

{\flushleft \textbf{Azimuth Patch Merging.}}
Corresponding to the polar coordinate-based patch embedding and self-attention, we design a polar azimuth patch merging.
The process of patch merging is similar to other window-based attention (e.g., Swin transformer~\cite{liu2021swin}), the difference is that our patch merged along the azimuth due to the radial strip window.
\section{Experiment}
In this section, firstly we introduce the experimental setting and details of implementation. 
Then, we compare our method based on six datasets: GoPro~\cite{nah2017deep}, HIDE~\cite{iccv2019_hide}, RealBlur~\cite{eccv2020real}, REDS~\cite{liu2022video}, RSBur~\cite{rim2022realistic}, and RWBI~\cite{zhang2020deblurring}.
Moreover, we analyze the effect of each module by different design choices in the ablation studies.
Because of the limitation of space, more results and implementation details are demonstrated in the supplementary materials.
\subsection{Exprimental Details}
\textbf{Metrics.}
We use PSNR and SSIM metrics to compare the performance for quantitative analysis.
To verify the efficiency of the model, we use parameters (M) to compare the performance with other methods, where the M stands for million.
All the validation of parameters and time is implemented on a single NVIDIA 3090 GPU, and tested on input images of $256 \times 256$ pixels.
The best and the second best results are \textbf{highlight} and \underline{underlined}.
\\
\textbf{Datasets.} We train our proposed RST on the GoPro dataset~\cite{nah2017deep}, which contains 2,103 pairs of blurry images for training and 1,111 pairs for testing. 
For generalization testing, we adopt more deblurring datasets.
The HIDE~\cite{iccv2019_hide} dataset includes 2,025 pairs of images mainly about humans.
Considering the real-world blurring scenes, we perform RST on the real-world datasets Realblur~\cite{eccv2020real} dataset, which contains 980 pairs of images for testing.
Another real-world dataset RWBI has 3,112 real-world images without ground truth.
We also evaluate our methods on REDS~\cite{liu2022video} and RSBlur~\cite{rim2022realistic} datasets.
\\
\textbf{Implementation Details.} 
We train our model with AdamW~\cite{kingma2014adam} optimier and the same loss function following ~\cite{cvpr2021cho}. 
Following the setting of \cite{kong2023efficient}, the initial learning rate is $1e^{-3}$ and decreases to $1e^{-7}$ after 300,000 iterations with the cosine decay strategy.
The training set of RST is 64 batch size and the $128 \times 128$ patch size on the GoPro~\cite{nah2017deep}, REDS~\cite{liu2022video}, and RSBlur~\cite{rim2022realistic} datasets.
Our model is based on 3-level encoder-decoder architecture, which has the number of Transformer blocks on [6, 6, 12].
The progressive learning strategy is adopted for our model by following the work~\cite{eccv2022_Stripformer}.
%
\begin{table}[!t]\footnotesize
\centering
\caption{Quantitative comparison on {GoPro}~\cite{eccv2020real} and {HIDE}~\cite{iccv2019_hide} for image debluring. 
All methods are only trained on GoPro~\cite{nah2017deep}.
}
\scalebox{0.900}{
\footnotesize
\setlength{\tabcolsep}{3mm}{
\begin{tabular}{cccccccccccc}
\toprule[0.8pt]
\multicolumn{2}{c}{\multirow{2}{*}{}} & 
\multicolumn{4}{c|}{\textbf{GoPro}~\cite{nah2017deep}} & \multicolumn{4}{c}{\textbf{HIDE}~\cite{iccv2019_hide}}& \multicolumn{2}{||c}{Params~$\downarrow$}  
\\ 
\multicolumn{2}{l|}{\textbf{Method}} & 
\multicolumn{2}{c}{PSNR~$\uparrow$}& \multicolumn{2}{c|}{SSIM~$\uparrow$} & \multicolumn{2}{c}{PSNR~$\uparrow$} & \multicolumn{2}{c}{SSIM~$\uparrow$}& \multicolumn{2}{||c}{(M)}
\\\midrule[0.8pt]

\multicolumn{2}{l|}{DeblurGAN-v2~\cite{kupyn2019deblurgan}}
& \multicolumn{2}{c}{29.55}& \multicolumn{2}{c|}{0.934}
& \multicolumn{2}{c}{27.40} & \multicolumn{2}{c|}{0.882}  
& \multicolumn{2}{||c}{60.9}  \\

\multicolumn{2}{l|}{DMPHN~\cite{zhang2019deep}} 
& \multicolumn{2}{c}{31.20}& \multicolumn{2}{c|}{0.940} 
& \multicolumn{2}{c}{29.09} & \multicolumn{2}{c|}{0.924}
& \multicolumn{2}{||c}{21.7}  \\

\multicolumn{2}{l|}{MIMO~\cite{cvpr2021cho}}  
& \multicolumn{2}{c}{32.45}& \multicolumn{2}{c|}{0.957} 
& \multicolumn{2}{c}{29.99} & \multicolumn{2}{c|}{0.930}
& \multicolumn{2}{||c}{{16.1}}  \\

\multicolumn{2}{l|}{MPRNet~\cite{zamir2021multi}}   
& \multicolumn{2}{c}{32.66}& \multicolumn{2}{c|}{0.959}
& \multicolumn{2}{c}{30.96} & \multicolumn{2}{c|}{0.939}
& \multicolumn{2}{||c}{20.1}  \\

\multicolumn{2}{l|}{Restormer~\cite{zamir2022restormer}}  
& \multicolumn{2}{c}{32.92}& \multicolumn{2}{c|}{0.961}
& \multicolumn{2}{c}{31.22} & \multicolumn{2}{c|}{0.942}
& \multicolumn{2}{||c}{26.1}  \\

\multicolumn{2}{l|}{Uformer~\cite{wang2022uformer}}  
& \multicolumn{2}{c}{33.06}& \multicolumn{2}{c|}{0.967}
& \multicolumn{2}{c}{30.90} & \multicolumn{2}{c|}{\textbf{0.953}} 
& \multicolumn{2}{||c}{50.9}  \\

\multicolumn{2}{l|}{Stripformer~\cite{eccv2022_Stripformer}}   
& \multicolumn{2}{c}{33.08} & \multicolumn{2}{c|}{0.962}
& \multicolumn{2}{c}{31.03} & \multicolumn{2}{c|}{0.940}
& \multicolumn{2}{||c}{19.7}  \\
\multicolumn{2}{l|}{SFNet~\cite{cui2022selective}}   
& \multicolumn{2}{c}{{33.27}} & \multicolumn{2}{c|}{0.962}
& \multicolumn{2}{c}{{31.10}} & \multicolumn{2}{c|}{0.941} 
& \multicolumn{2}{||c}{\textbf{13.3}}  \\
\multicolumn{2}{l|}{FFTformer~\cite{kong2023efficient}}   
& \multicolumn{2}{c}{\underline{34.21}} & \multicolumn{2}{c|}{\textbf{0.969}}
& \multicolumn{2}{c}{{31.62}} & \multicolumn{2}{c|}{0.946} 
& \multicolumn{2}{||c}{16.6}  \\

\multicolumn{2}{l|}{UFPNet~\cite{fang2023self}}   
& \multicolumn{2}{c}{34.06} & \multicolumn{2}{c|}{\underline{0.968}}
 & \multicolumn{2}{c}{\underline{31.74}} & \multicolumn{2}{c|}{0.947}
& \multicolumn{2}{||c}{80.3}  \\

 \midrule
\multicolumn{2}{l|}{\text{RST (Ours)}} 
& \multicolumn{2}{c}{\textbf{34.25}} & \multicolumn{2}{c|}{\textbf{0.969}}
& \multicolumn{2}{c}{\textbf{31.84}} & \multicolumn{2}{c|}{\underline{0.948}}
& \multicolumn{2}{||c}{\underline{14.3}}  
  \\
\bottomrule[0.8pt]
\end{tabular}}
}
\label{tab:deblur_goprohide}
\end{table}
\subsection{Experimental Results}
{\flushleft \textbf{Quantitative Analysis.}}
We compare our model on GoPro~\cite{nah2017deep} testing set and HIDE~\cite{iccv2019_hide} dataset with 10 SOTA methods. 
Table~\ref{tab:deblur_goprohide} shows that the proposed RST exhibits superior performance over other methods on two benchmarks.
For the GoPro test set, it is worth mentioning that compared with Stripformer~\cite{eccv2022_Stripformer}, which was designed based on the Cartesian system, RST achieves a performance gain of 1.17dB in PSNR.
We also compare RST on the HIDE dataset.
Notably, our proposed method achieves the best performance by only training on the GoPro training set, which demonstrates stronger generalization capability than the best previous methods FFTformer~\cite{kong2023efficient}.
Compared with FFTformer, RST has a performance boost of 0.22 dB with PSNR metrics.
Furthermore, Table~\ref{tab:deblur_real} reports more results on the RealBlur dataset, and RST achieves the second-best performance on RealBlur-J and the best on RealBlur-R.
It is significant to mention that RST obtains a performance gain of 0.67dB over the FFTformer on average of the RealBlur dataset. 
%
\begin{table}[t]\footnotesize
\centering
\caption{Quantitative comparison on {RealBlur}~\cite{eccv2020real}. 
All methods are only trained on GoPro~\cite{nah2017deep}.
}
\scalebox{0.900}{
\footnotesize
\setlength{\tabcolsep}{2mm}{
\begin{tabular}{cccccccccccccc}
\toprule[0.8pt]
\multicolumn{2}{c}{\multirow{2}{*}{}} & 
\multicolumn{4}{c|}{\textbf{RealBlur-R}~\cite{eccv2020real}} & \multicolumn{4}{c}{\textbf{RealBlur-J}~\cite{eccv2020real}}& \multicolumn{4}{||c}{Average}  
\\ 
\multicolumn{2}{l|}{\textbf{Method}} & 
\multicolumn{2}{c}{PSNR~$\uparrow$}& \multicolumn{2}{c|}{SSIM~$\uparrow$} & \multicolumn{2}{c}{PSNR~$\uparrow$} & \multicolumn{2}{c}{SSIM~$\uparrow$}& \multicolumn{2}{||c}{PSNR~$\uparrow$} & \multicolumn{2}{c}{SSIM~$\uparrow$} 
\\\midrule[0.8pt]

\multicolumn{2}{l|}{IR-SDE~
\cite{luo2023image}}& \multicolumn{2}{c}{32.56} & \multicolumn{2}{c|}{0.909} & \multicolumn{2}{c}{23.19}  & \multicolumn{2}{c}{0.691} & \multicolumn{2}{||c}{27.89} & \multicolumn{2}{c}{0.800} \\
\multicolumn{2}{l|}{NAFNet~\cite{chen2022simple}}
  &  \multicolumn{2}{c}{33.63} & \multicolumn{2}{c|}{\underline{0.944}} & \multicolumn{2}{c}{26.33} & \multicolumn{2}{c}{\textbf{0.856}}& \multicolumn{2}{||c}{29.98} & \multicolumn{2}{c}{\underline{0.900}}  
  \\
\multicolumn{2}{l|}{FFTformer~\cite{kong2023efficient}}
  &  \multicolumn{2}{c}{33.66} & \multicolumn{2}{c|}{\textbf{0.948}} & \multicolumn{2}{c}{25.71} & \multicolumn{2}{c}{0.851}& \multicolumn{2}{||c}{29.69} & \multicolumn{2}{c}{\underline{0.900}}  
  \\

\multicolumn{2}{l|}{CODE~
 \cite{Zhao_2023_CVPR}}                     & \multicolumn{2}{c}{33.81}& \multicolumn{2}{c|}{0.939} & \multicolumn{2}{c}{26.25} & \multicolumn{2}{c|}{0.801}  & \multicolumn{2}{||c}{30.03} & \multicolumn{2}{c}{0.870}\\
\multicolumn{2}{l|}{GRL-B~\cite{li2023grl}}                
  & \multicolumn{2}{c}{\underline{33.97}} & \multicolumn{2}{c|}{\underline{0.944}} & \multicolumn{2}{c}{{\textbf{26.40}}} & \multicolumn{2}{c}{0.816}& \multicolumn{2}{||c}{\underline{30.19}} & \multicolumn{2}{c}{0.880}  
  \\
 \midrule
  \multicolumn{2}{l|}{{}{RST} (Ours)} 
  &  \multicolumn{2}{c}{\textbf{34.37}} & \multicolumn{2}{c|}{\textbf{0.948}} & \multicolumn{2}{c}{\underline{26.35}} & \multicolumn{2}{c}{\underline{0.855}}  & \multicolumn{2}{||c}{\textbf{30.36}} & \multicolumn{2}{c}{\textbf{0.902}}  
  \\
\bottomrule[0.8pt]
\end{tabular}}
}
\label{tab:deblur_real}
\end{table}
\begin{table}[t]\footnotesize
\centering
\caption{Quantitative comparison on {RWBI}~\cite{zhang2020deblurring} with the no-reference metric NIQE.
}
\footnotesize
\scalebox{0.900}{
\setlength{\tabcolsep}{1mm}
{
\begin{tabular}{cccccccccccccccccccc}
\toprule[0.8pt]
\multicolumn{2}{l|}{Method} 
& \multicolumn{2}{c}{Input} 
& \multicolumn{2}{c}{Uformer~\cite{wang2022uformer}}
& \multicolumn{2}{c}{SFNet~\cite{cui2022selective}}
& \multicolumn{2}{c}{FFTformer~\cite{kong2023efficient}}
& \multicolumn{2}{c|}{UFPNet~\cite{fang2023self}}
& \multicolumn{2}{c}{RST (Ours)}
\\ 
\midrule[0.8pt]
\multicolumn{2}{l|}{NIQE$\downarrow$} 
 & \multicolumn{2}{c}{5.436} 
& \multicolumn{2}{c}{6.061}
& \multicolumn{2}{c}{5.683}
& \multicolumn{2}{c}{5.188}
& \multicolumn{2}{c|}{5.467} 
& \multicolumn{2}{c}{\textbf{4.929}}\\
\bottomrule[0.8pt]
\end{tabular}
}
}
\label{table4rwbi}
\end{table}
Besides the synthesis datasets, we further conduct experiments on more real-world datasets to evaluate the capability of our method on real-world image deblurring. 
We train the RST on a newly proposed real-world dataset RSBlur~\cite{rim2022realistic}, which contains 8,878, 1,120, and 3,360 blurred images of train, validation, and test sets respectively, following the training strategies of SFNet~\cite{cui2022selective}.
We compare our model on the RSBlur testing set with 6 SOTA methods: SRN~\cite{kupyn2019deblurgan}, MIMO~\cite{cvpr2021cho}, MPRNet~\cite{zamir2021multi}, Restormer~\cite{zamir2022restormer}, Uformer~\cite{wang2022uformer}, FFTformer~\cite{kong2023efficient}.
As shown in Table~\ref{tab:rsblur}, RST achieves a new state-of-the-art and makes a clear gain of 0.40dB, and has a performance of 0.43dB compared with FFTformer~\cite{kong2023efficient}.
We perform our method on REDS~\cite{liu2022video} datsets, and following the training set of ~\cite{chen2022simple}.
We conduct the evaluation experiments on REDS-val-300~\cite{nah2021ntire}, which contain 300 images from the validation set following~\cite{chen2022simple}.
Compared with FFTformer~\cite{kong2023efficient}, the improvement is 0.11 dB in terms of PSNR.
As shown in Table~\ref{tab:reds}, RST outperforms all the compared methods and achieves 0.16 dB and 0.42 dB improvement over the existing best-performance methods NAFNet~\cite{chen2022simple} and the winning solution of NTIRE 2021~\cite{nah2021ntire} HINet~\cite{chen2021hinet}.
Furthermore, we conduct RST on RWBI dataset~\cite{zhang2020deblurring}, which contains 3,112 real-world images without ground truth.
As shown in Table~\ref{table4rwbi}, RST has lower NIQE scores compared with other SOTA works, which means the superior performance of perceptual quality.
%

\begin{figure}[t]
\scriptsize
\centering
\begin{tabular}{ccc}
\begin{adjustbox}{valign=t}
\begin{tabular}{cccc}
\includegraphics[width=0.230\textwidth]{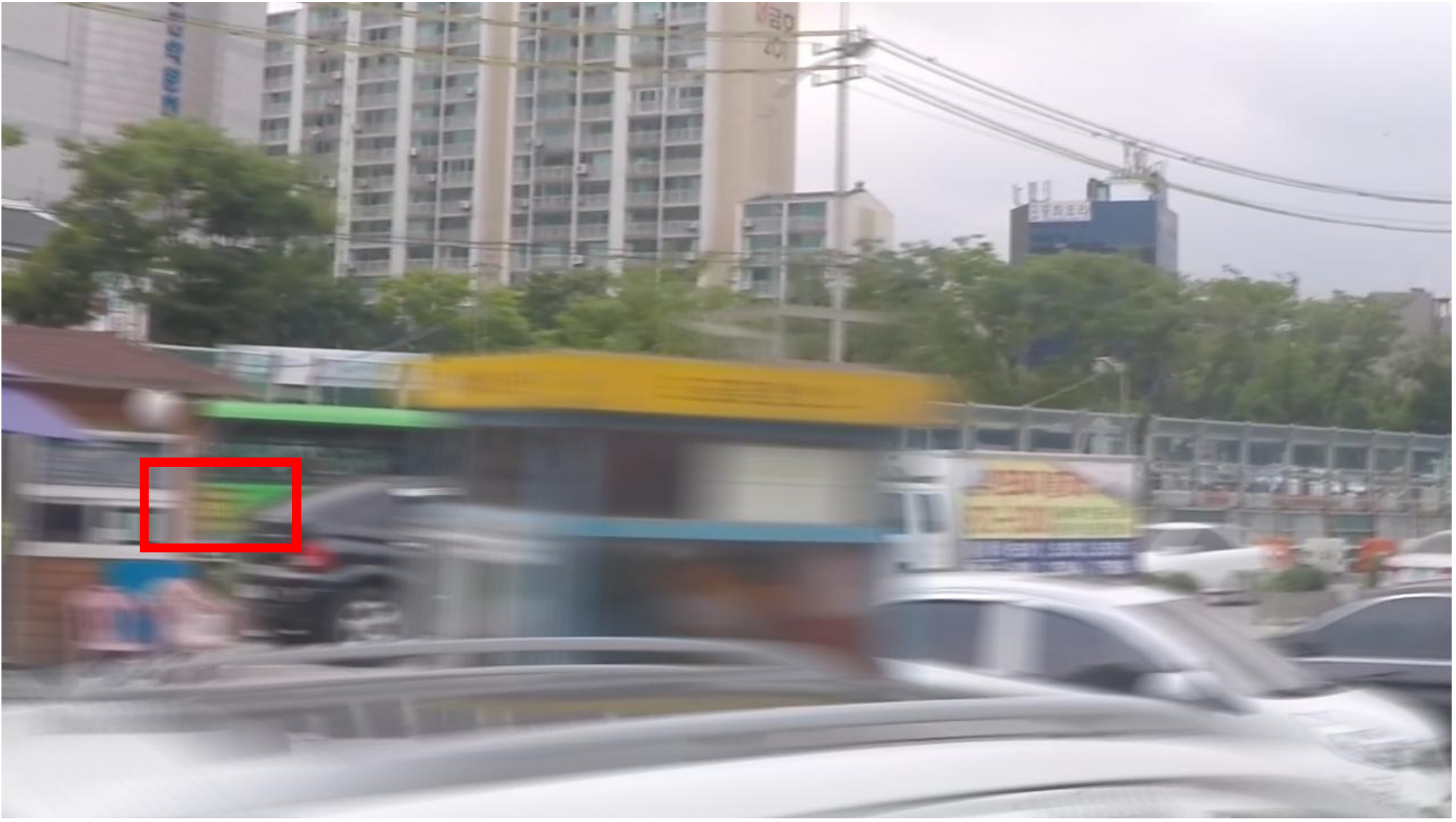} \hspace{0mm} &
\includegraphics[width=0.230\textwidth]{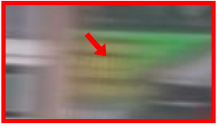} \hspace{0mm} &
\includegraphics[width=0.230\textwidth]{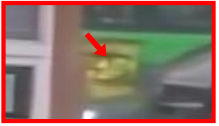} \hspace{0mm} &
\includegraphics[width=0.230\textwidth]{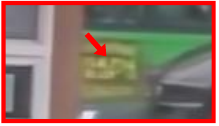} \hspace{0mm}  \\
Blurry Image\hspace{0mm} &
Blurry \hspace{0mm} &
MPRNet~\cite{zamir2021multi} \hspace{0mm} &
Uformer~\cite{wang2022uformer} \hspace{0mm} \\

\includegraphics[width=0.230\textwidth]{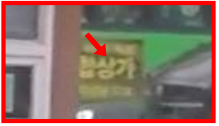} \hspace{0mm} &
\includegraphics[width=0.230\textwidth]{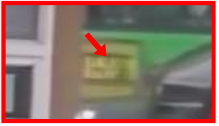} \hspace{0mm} &
\includegraphics[width=0.230\textwidth]{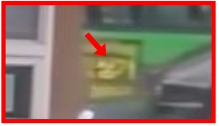} \hspace{0mm} &
\includegraphics[width=0.230\textwidth]{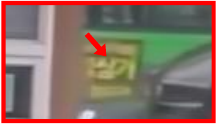} \hspace{0mm}  \\
Reference \hspace{-4mm} &
Stripformer~\cite{eccv2022_Stripformer} \hspace{0mm} &
FFTformer~\cite{kong2023efficient} \hspace{0mm} &
\textbf{RST (Ours)} \hspace{0mm} \\

\end{tabular}
\end{adjustbox}
\end{tabular}
\vspace{-2mm}
\caption{The qualitative results on GoPro~\cite{nah2017deep}. 
The proposed method achieves the clearest result of restoring the characters on the green bus.}
\label{pic:gopro_pic}
\end{figure}

\begin{figure}[t]
\scriptsize
\centering
\begin{tabular}{ccc}
\begin{adjustbox}{valign=t}
\begin{tabular}{cccc}
\includegraphics[width=0.230\textwidth]{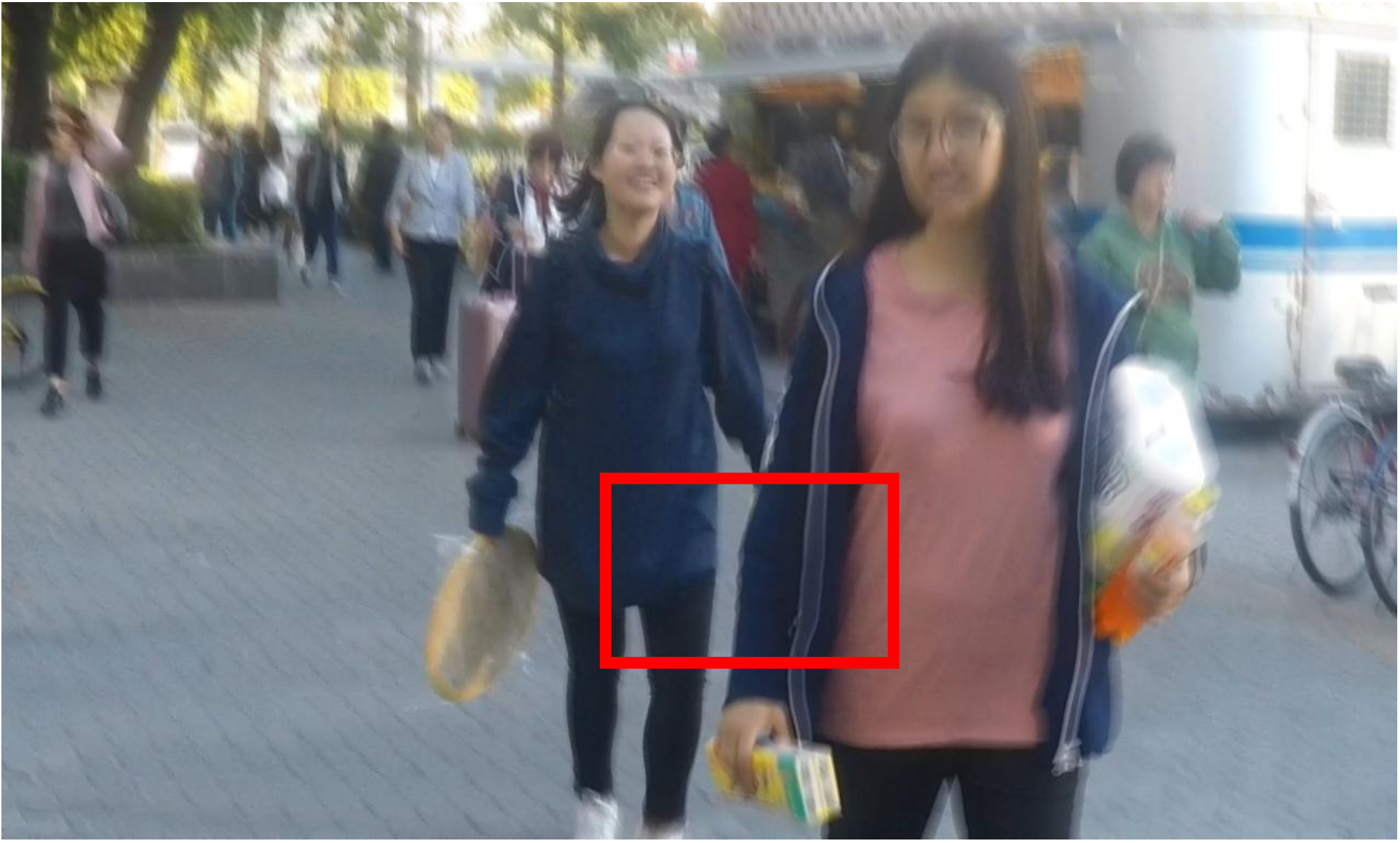} \hspace{0mm} &
\includegraphics[width=0.230\textwidth]{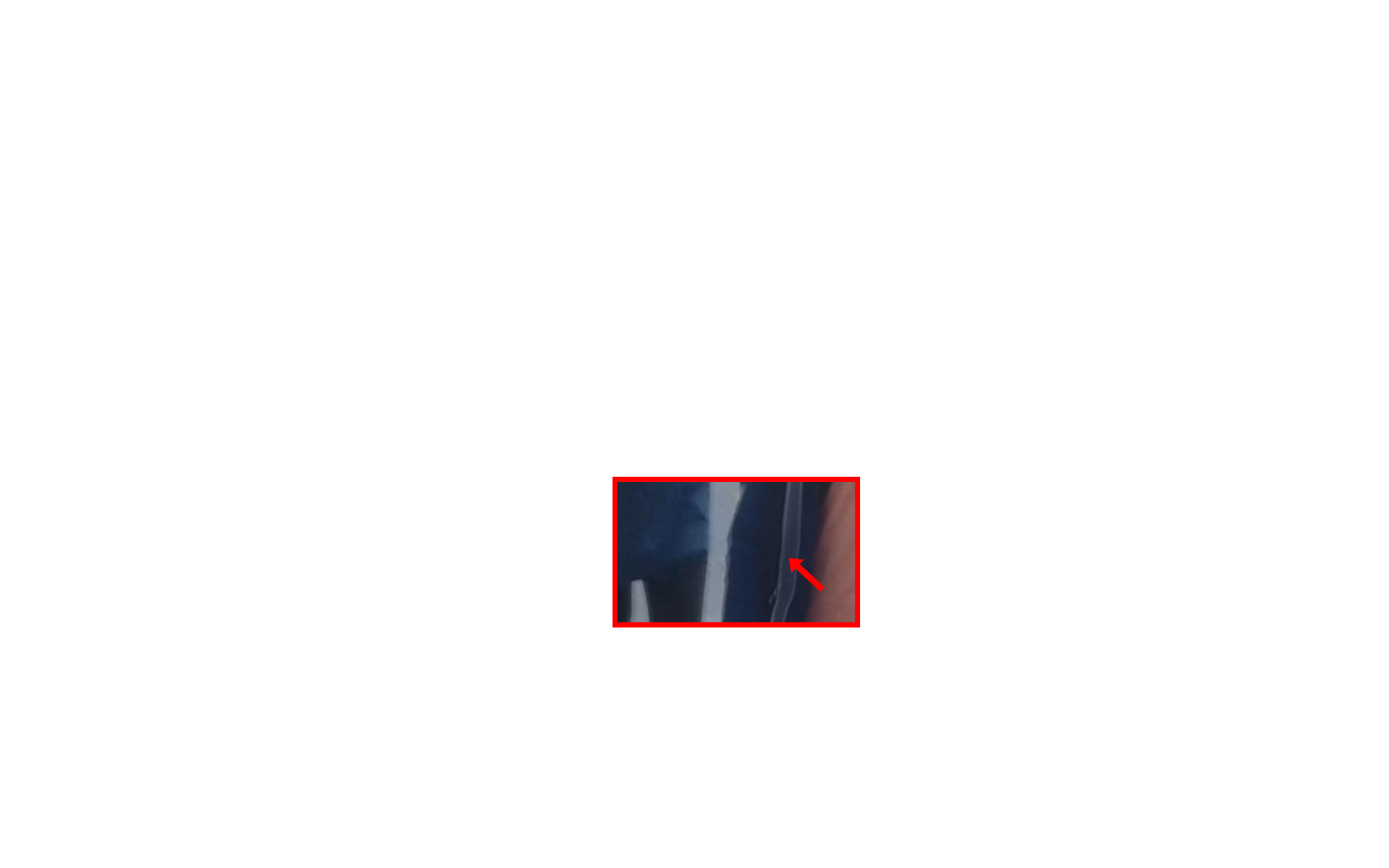} \hspace{0mm} &
\includegraphics[width=0.230\textwidth]{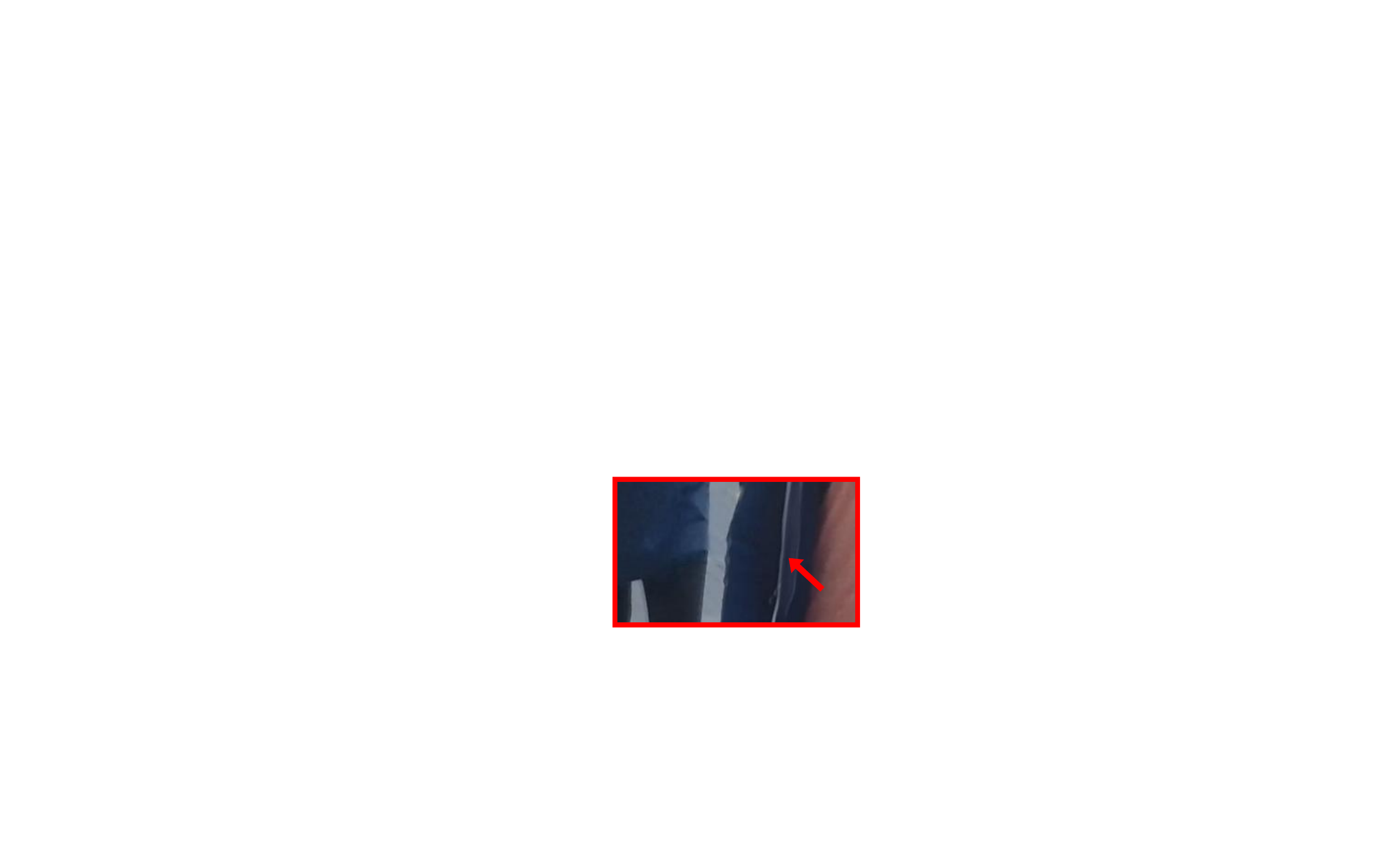} \hspace{0mm} &
\includegraphics[width=0.230\textwidth]{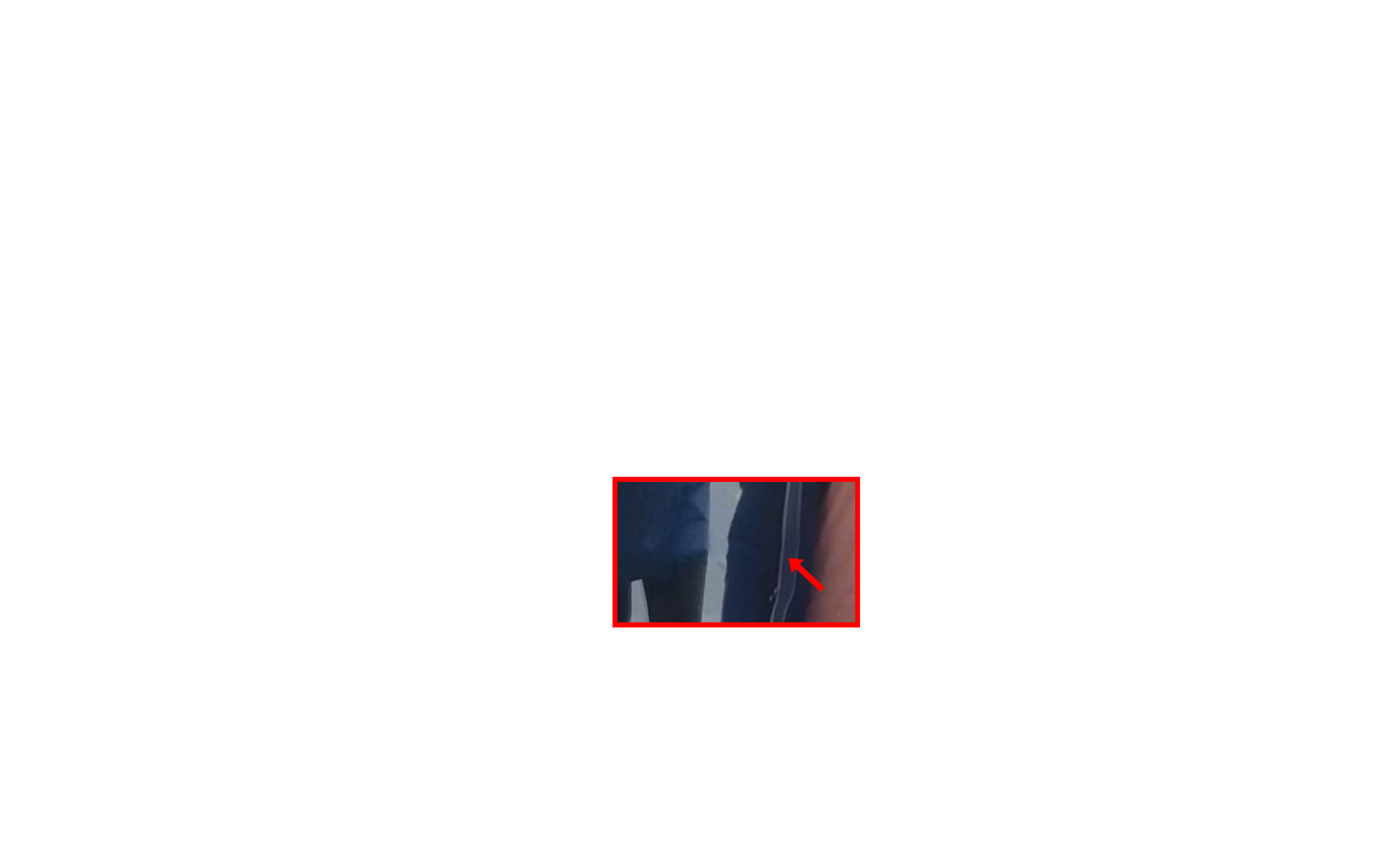} \hspace{0mm}  \\
Blurry Image\hspace{0mm} &
Blurry \hspace{0mm} &
Restormer~\cite{zamir2022restormer} \hspace{0mm} &
Uformer~\cite{wang2022uformer} \hspace{0mm} \\

\includegraphics[width=0.230\textwidth]{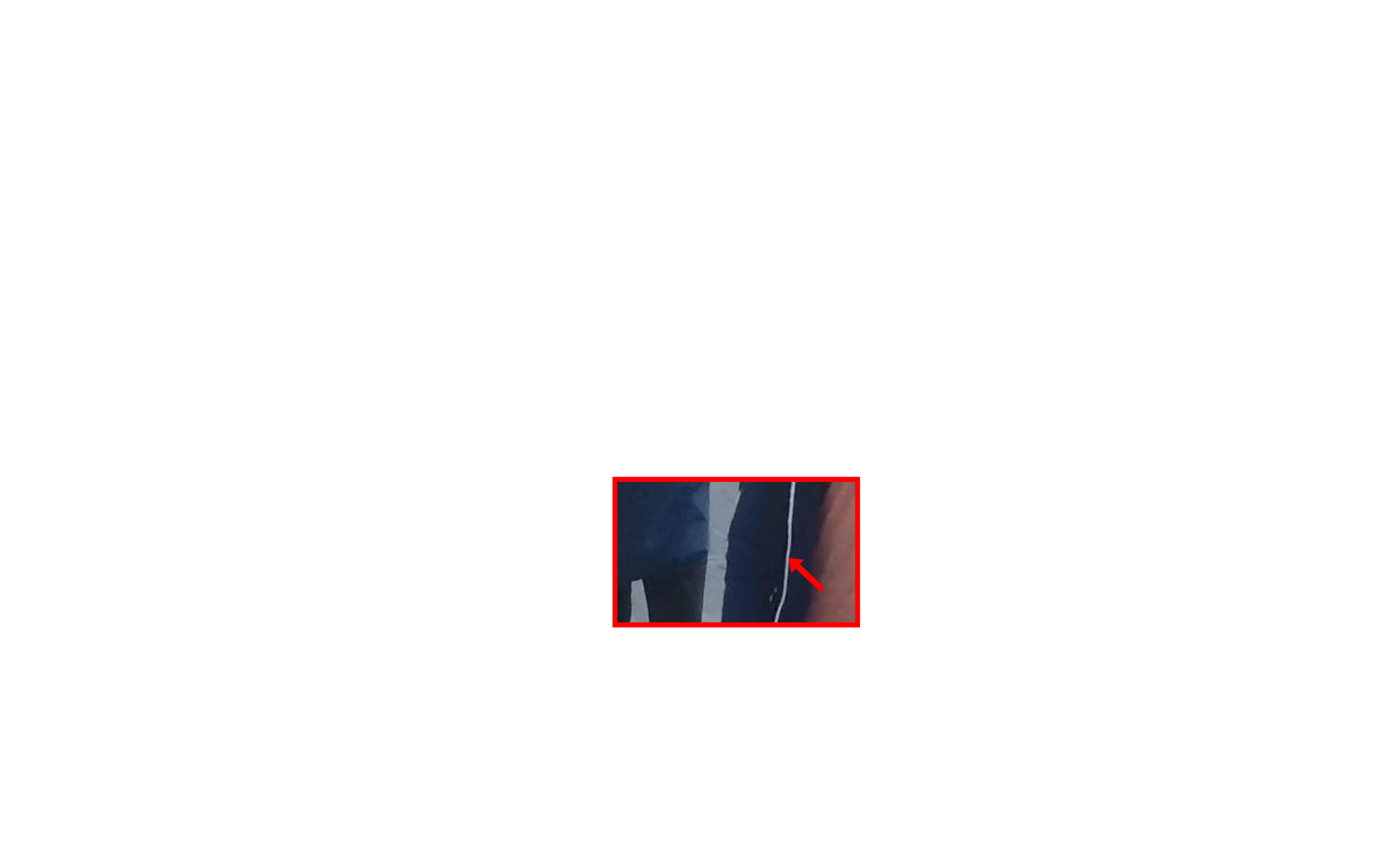} \hspace{0mm} &
\includegraphics[width=0.230\textwidth]{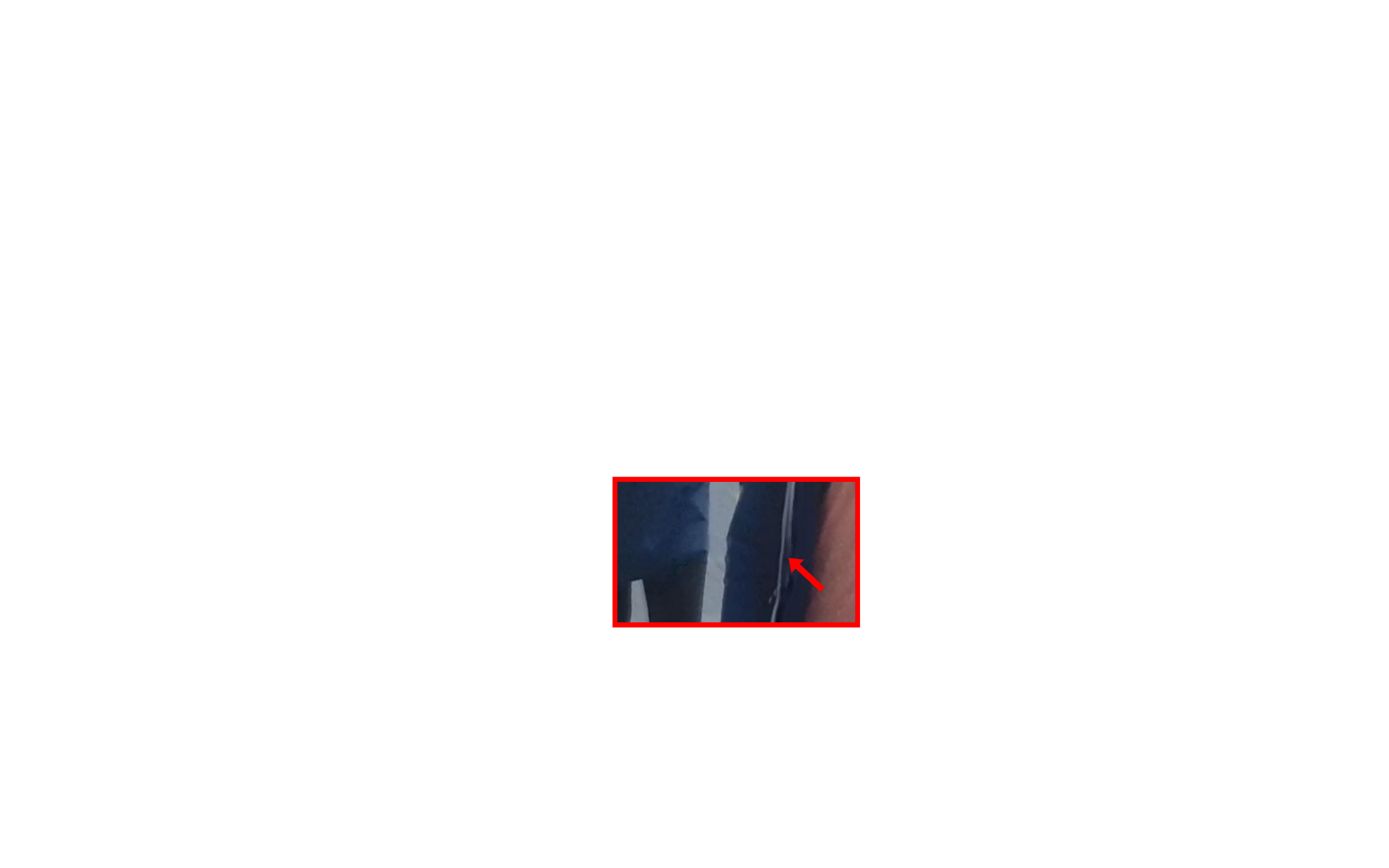} \hspace{0mm} &
\includegraphics[width=0.230\textwidth]{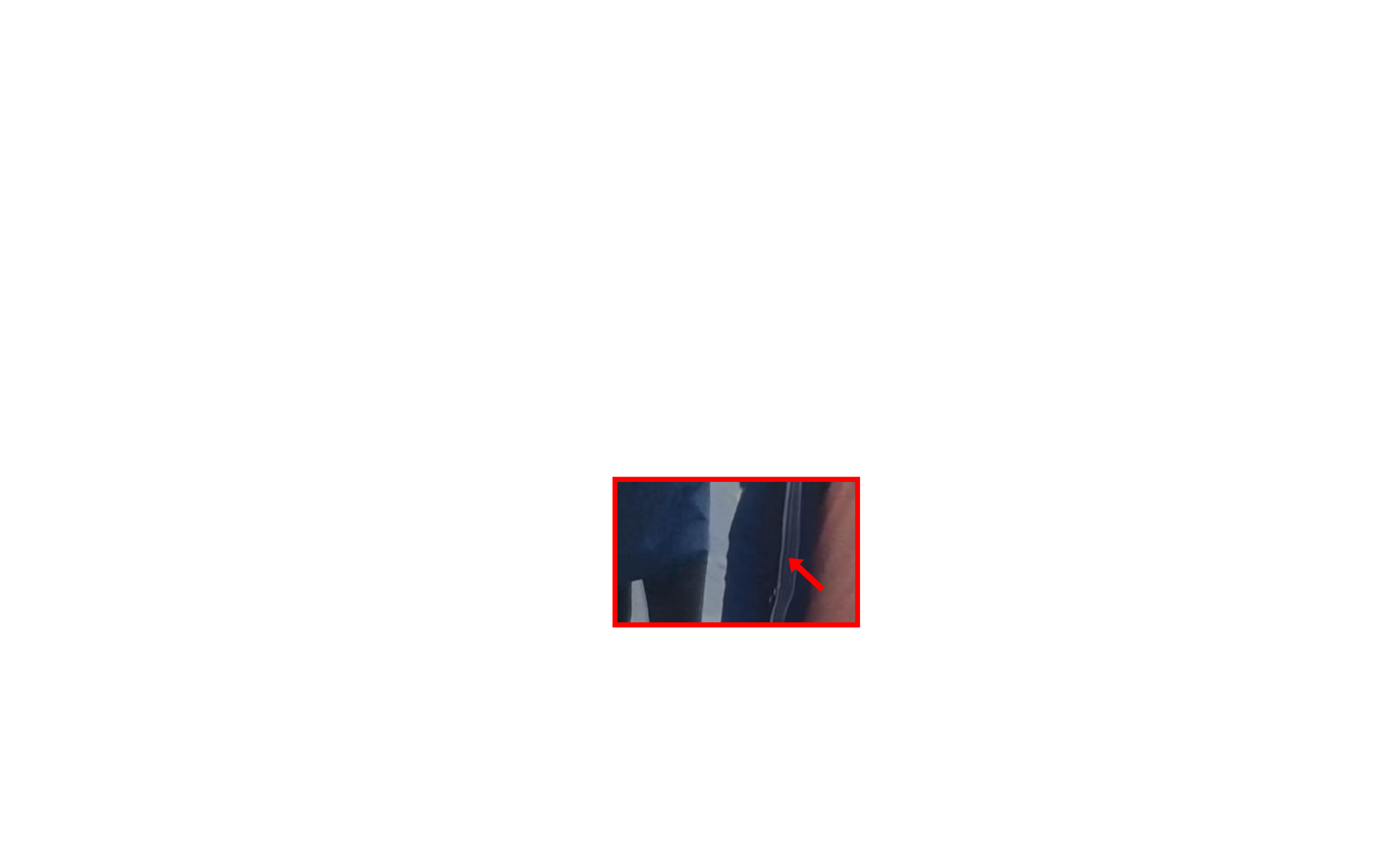} \hspace{0mm} &
\includegraphics[width=0.230\textwidth]{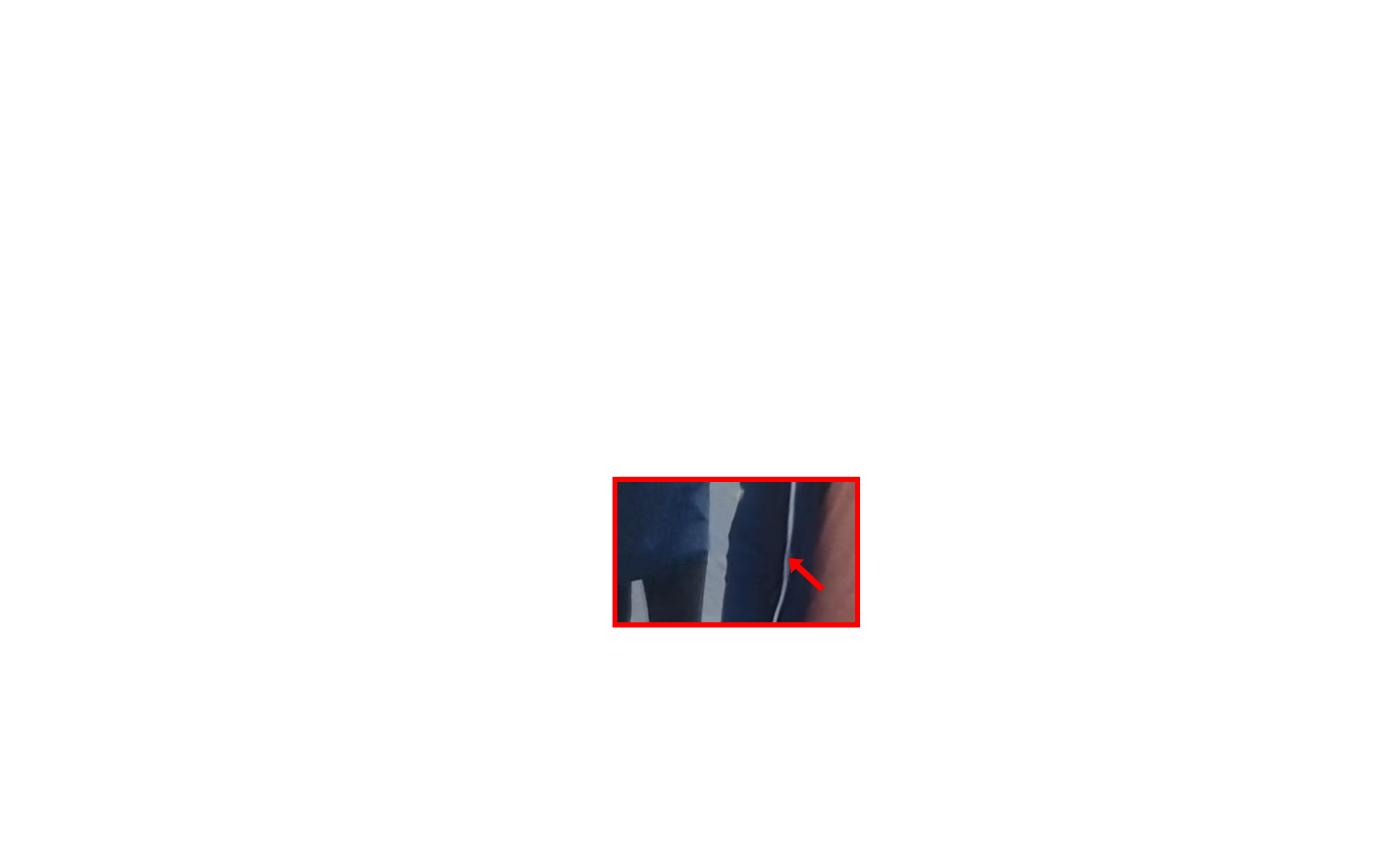} \hspace{0mm}  \\
Reference \hspace{0mm} &
Stripformer~\cite{eccv2022_Stripformer} \hspace{0mm} &
FFTformer~\cite{kong2023efficient} \hspace{0mm} &
\textbf{RST (Ours)} \hspace{0mm} \\
\end{tabular}
\end{adjustbox}
\end{tabular}
\caption{The qualitative results on HIDE~\cite{iccv2019_hide}. The proposed method restores a clearer result on the structure of the women's zipper.
}
\label{pic:hide_pic}
\end{figure}

\begin{figure}[t]
\scriptsize
\centering
\begin{tabular}{ccc}
\begin{adjustbox}{valign=t}
\begin{tabular}{c}
\includegraphics[width=0.20\textwidth]{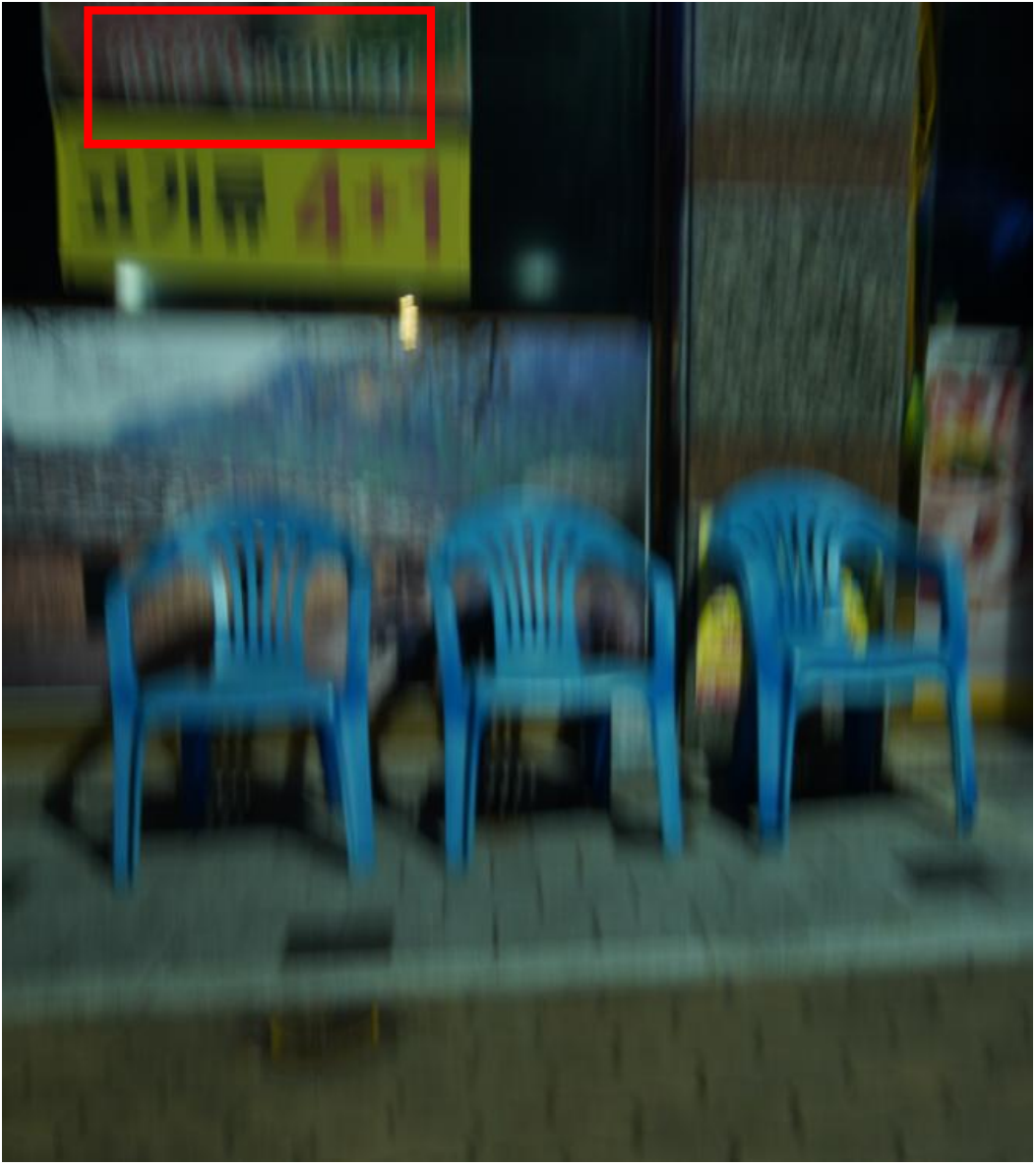}
\\
Blurry Image
\end{tabular}
\end{adjustbox}
\hspace{-1mm}
\begin{adjustbox}{valign=t}
\begin{tabular}{ccc}
\includegraphics[width=0.24\textwidth]{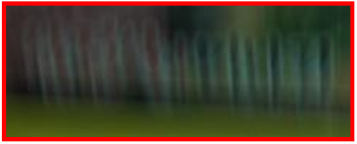} \hspace{-1mm} &
\includegraphics[width=0.24\textwidth]{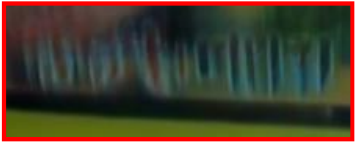} \hspace{-1mm} &
\includegraphics[width=0.24\textwidth]{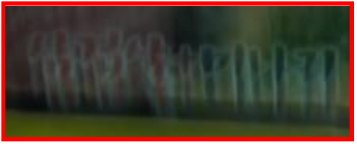} \hspace{-1mm} 
\\
Blurry \hspace{-1mm} &
NAFNet~\cite{chen2022simple} \hspace{-1mm} &
CODE~\cite{Zhao_2023_CVPR} \hspace{-1mm} 
\\
\includegraphics[width=0.24\textwidth]{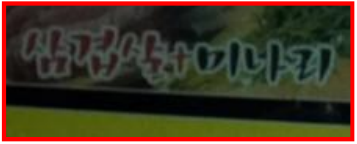} \hspace{-1mm} &
\includegraphics[width=0.24\textwidth]{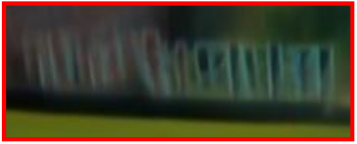} \hspace{-1mm} &
\includegraphics[width=0.24\textwidth]{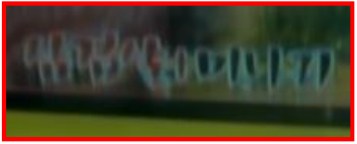} \hspace{-1mm}
\\ 
Reference \hspace{-1mm} &
FFTformer~\cite{kong2023efficient} \hspace{-1mm} &
\textbf{RST (Ours)} \hspace{-1mm} 
\\
\end{tabular}
\end{adjustbox}
\end{tabular}
\caption{The qualitative results on RealBlur~\cite{eccv2020real}, The proposed method generates much clearer content for the advertising board. }
\label{pic:realblur}
\end{figure}

{\flushleft \textbf{Qualitative Analysis.}}
We perform the qualitative comparisons on the GoPro, HIDE, and RealBlur in Figure~\ref{pic:gopro_pic}, Figure~\ref{pic:hide_pic}, and Figure~\ref{pic:realblur}, respectively.
Compared with other SOTA methods, our proposed approach restores clearer results. 
For example the sharper characters on the green bus (Figure~\ref{pic:gopro_pic}), the finer edge of the zipper (Figure~\ref{pic:hide_pic}), and the much clearer text of the advertising board (Figure~\ref{pic:realblur}).
%
%
\begin{table*}[!t]
\begin{minipage}{.475\linewidth}
\caption{Quantitative comparison on {RSBlur}~\cite{rim2022realistic}. 
All methods are only trained on RSBlur~\cite{rim2022realistic}.
}
\centering
\scalebox{0.900}{
\footnotesize
\setlength{\tabcolsep}{0mm}
{
\begin{tabular}{cccccccccc}

\toprule[0.8pt]
\multicolumn{2}{l|}{\textbf{Method}} &\multicolumn{2}{c|}{PSNR~$\uparrow$}& \multicolumn{2}{c|}{~SSIM~$\uparrow$} & \multicolumn{2}{c}{Params~$\downarrow$} \\ 

\midrule[0.8pt]
\multicolumn{2}{l|}{SRN~\cite{tao2018scale}}           
& \multicolumn{2}{c|}{32.53} & \multicolumn{2}{c|}{0.840}& \multicolumn{2}{c}{\textbf{6.8}} \\
\multicolumn{2}{l|}{MIMO~\cite{cvpr2021cho}}           
& \multicolumn{2}{c|}{33.37} & \multicolumn{2}{c|}{0.856}& \multicolumn{2}{c}{16.1}  \\
\multicolumn{2}{l|}{MPRNet~\cite{zamir2021multi}}           
& \multicolumn{2}{c|}{33.61} & \multicolumn{2}{c|}{0.861}& \multicolumn{2}{c}{20.1} \\
\multicolumn{2}{l|}{Restormer~\cite{zamir2022restormer}}           
& \multicolumn{2}{c|}{33.69} & \multicolumn{2}{c|}{0.863}& \multicolumn{2}{c}{26.1}  \\
\multicolumn{2}{l|}{FFTformer~\cite{kong2023efficient}}           
& \multicolumn{2}{c|}{33.95} & \multicolumn{2}{c|}{\underline{0.894}}& \multicolumn{2}{c}{16.6}  \\
\multicolumn{2}{l|}{Uformer~\cite{wang2022uformer}}           
& \multicolumn{2}{c|}{\underline{33.98}} & \multicolumn{2}{c|}{0.866}& \multicolumn{2}{c}{50.9} \\
\midrule
\multicolumn{2}{l|}{\text{RST (Ours)}} 
& \multicolumn{2}{c|}{\textbf{34.38}} & \multicolumn{2}{c|}{\textbf{0.899}}& \multicolumn{2}{c}{\underline{14.3}} \\

\bottomrule[0.8pt]
\end{tabular}}
}
\label{tab:rsblur}
\vspace{-3.7mm}
\end{minipage}~~~~~\begin{minipage}{.48\linewidth}
\caption{Quantitative comparison on {REDS-val-300}~\cite{nah2021ntire}. 
All methods are only trained on REDS~\cite{liu2022video}.
}
\centering
\scalebox{0.900}{
\footnotesize
\setlength{\tabcolsep}{0mm}
{
\begin{tabular}{cccccccccc}
\toprule[0.8pt]
\multicolumn{2}{l|}{\textbf{Method}} &\multicolumn{2}{c|}{PSNR~$\uparrow$}& \multicolumn{2}{c|}{~SSIM~$\uparrow$} & \multicolumn{2}{c}{Params~$\downarrow$}  \\ 
\midrule[0.8pt]
\multicolumn{2}{l|}{MPRNet~\cite{zamir2021multi}}           
& \multicolumn{2}{c|}{28.79} & \multicolumn{2}{c|}{0.811}& \multicolumn{2}{c}{20.1}  \\
\multicolumn{2}{l|}{HINet~\cite{chen2021hinet}}           
& \multicolumn{2}{c|}{28.83} & \multicolumn{2}{c|}{0.862}& \multicolumn{2}{c}{88.7} \\
\multicolumn{2}{l|}{MAXIM~\cite{tu2022maxim}}           
& \multicolumn{2}{c|}{28.93} & \multicolumn{2}{c|}{0.865}& \multicolumn{2}{c}{22.2}  \\
\multicolumn{2}{l|}{NAFNet~\cite{chen2022simple}}           
& \multicolumn{2}{c|}{29.09} & \multicolumn{2}{c|}{\underline{0.867}}& \multicolumn{2}{c}{67.9}  \\
\multicolumn{2}{l|}{FFTformer~\cite{kong2023efficient}}           
& \multicolumn{2}{c|}{\underline{29.14}} & \multicolumn{2}{c|}{\underline{0.867}}& \multicolumn{2}{c}{\underline{16.6}} \\
\midrule

\multicolumn{2}{l|}{\text{RST (Ours)}} 
& \multicolumn{2}{c|}{\textbf{{29.25}}} & \multicolumn{2}{c|}{\textbf{{0.869}}}
& \multicolumn{2}{c}{\textbf{14.3}}\\

\bottomrule[0.8pt]
\end{tabular}}
}
\label{tab:reds}
\end{minipage}
\end{table*}
\begin{table*}[t]
\begin{minipage}{.40\linewidth}
\caption{Evaluations of DRE with different sector numbers on GoPro~\cite{nah2017deep} dataset.  
}
\centering
\scalebox{0.900}{
\footnotesize
\setlength{\tabcolsep}{2mm}
{
\begin{tabular}{cccccccc}
\toprule[0.8pt]
\multicolumn{1}{c}{{Mask Parts}}  &  \multicolumn{2}{c}{{PSNR$\uparrow$}} & \multicolumn{2}{c}{{SSIM$\uparrow$}}
\\ \midrule[0.8pt]
\multicolumn{1}{c}{2} 
&  \multicolumn{2}{c}{31.15} &  \multicolumn{2}{c}{0.9463} \\
\multicolumn{1}{c}{4} 
&  \multicolumn{2}{c}{31.17}&  \multicolumn{2}{c}{0.9461}  \\

\multicolumn{1}{c}{8}  
&  \multicolumn{2}{c}{31.14} &  \multicolumn{2}{c}{0.9464} \\

\multicolumn{1}{c}{16}  
&  \multicolumn{2}{c}{31.01} &  \multicolumn{2}{c}{0.9453} \\

\bottomrule
\end{tabular}}
}
\label{tab:abl-mask}
\vspace{-2.5mm}
\end{minipage}~~~~~\begin{minipage}{.54\linewidth}
\caption{Quantitative evaluations of each module in proposed method on the GoPro~\cite{nah2017deep} dataset. The number of sector is 4.}
\centering
\scalebox{0.900}{
\footnotesize
\setlength{\tabcolsep}{1mm}
{
\begin{tabular}{ccccccccccc}
\toprule[0.8pt]
 & \multicolumn{2}{c}{{DRE}} & \multicolumn{2}{c}{{RSAS}} & \multicolumn{2}{|c}{{Params $\downarrow$}} & \multicolumn{2}{c}{{FLOPs $\downarrow$}} &  \multicolumn{2}{c}{{PSNR $\uparrow$}} 
\\ \midrule[0.8pt]
& \multicolumn{2}{c}{\XSolidBrush} &\multicolumn{2}{c}{\XSolidBrush} 
&  \multicolumn{2}{|c}{16.5605} &  \multicolumn{2}{c}{131.75} &  \multicolumn{2}{c}{29.15}\\

& \multicolumn{2}{c}{\XSolidBrush} &\multicolumn{2}{c}{\CheckmarkBold } 
&  \multicolumn{2}{|c}{14.3046} &  \multicolumn{2}{c}{112.51} &  \multicolumn{2}{c}{31.00}\\
  
& \multicolumn{2}{c}{\CheckmarkBold } &\multicolumn{2}{c}{\CheckmarkBold } 
&  \multicolumn{2}{|c}{14.3041} &  \multicolumn{2}{c}{112.42} &  \multicolumn{2}{c}{31.01}\\

\bottomrule
\end{tabular}}
}
\label{tab:abl-module}
\end{minipage}
\end{table*}
\subsection{Ablation Studies}
\begin{figure*}[!t]
\begin{minipage}{.475\linewidth}
\vspace{-3.5mm}
\tiny
\centering
\begin{adjustbox}{valign=t}
\begin{tabular}{ccc}
\includegraphics[width=0.3\textwidth]{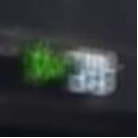} \hspace{0mm} &
\includegraphics[width=0.3\textwidth]{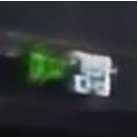} \hspace{0mm} &
\includegraphics[width=0.3\textwidth]{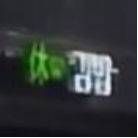} \hspace{0mm}  \\
(a)Blurry image \hspace{0mm} &
(b)w/o RSAS \hspace{0mm} &
(c)w/ RSAS \hspace{0mm} 
\\
\end{tabular}
\end{adjustbox}
\caption{
The comparison of RSAS. Adopting the RSAS module performs a better performance on image deblurring.
}
\label{pic:abl_fft}
\end{minipage}~~~~\begin{minipage}{.47\linewidth}
\tiny
\centering
\hspace{0cm}
\begin{adjustbox}{valign=t}
\begin{tabular}{ccc}
\hspace{-5mm}
\begin{adjustbox}{valign=t}
\begin{tabular}{ccc}
\includegraphics[width=0.3\textwidth]{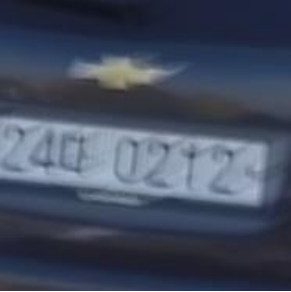} \hspace{-2mm} &
\includegraphics[width=0.3\textwidth]{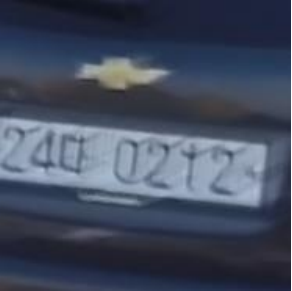} \hspace{-2mm} &
\includegraphics[width=0.3\textwidth]{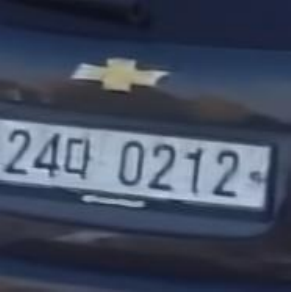} \hspace{-2mm}  \\
(a)Blurry image \hspace{-1mm} &
(b)Cartesian system \hspace{-1mm} &
(c)Polar system \hspace{-1mm} 
\\
\end{tabular}
\end{adjustbox}
\end{tabular}
\end{adjustbox}
\caption{
The comparison of different coordinate systems. RST restores a clearer image under the polar system than the Cartesian one.
} 
\label{pic:abl_swin}
\end{minipage}
\vspace{-5mm}
\end{figure*}

In this section, we analyze our proposed method with computation consumption and verify the effect of each component. 
All the ablation studies are trained with the GoPro~\cite{nah2017deep} dataset with 150,000 iterations for fair comparisons.
The parameters are calculated by an input tensor with the size of 256 $\times$ 256.\\
\textbf{Component Analysis.} 
Our RST contains the DRE module and the RSAS to obtain the shallow features and deep features to address various blur patterns of any direction.
Table~\ref{tab:abl-module} shows each module's contributions to RST, DRE denotes the dynamic radial embedding module and RSAS represents the radial strip attention solver.
\\
\textbf{Effect of DRE.}
We proposed the DRE to extract the shallow features from a polar coordinate way.
As the angle in the polar system, the sector number of the polar mask layer is also an important setting to analyze.
Table~\ref{tab:abl-mask} shows the performance of different sets of sector numbers.
With the partition number increasing, memory usage and computing consumption do not have a pronounced change.
The last two rows show that excessive fine-grained partitioning does not lead to better performance.
That is probably because splitting too many sectors will lead the detail information loss in the shallow feature extraction.
In the DRE module, we set the number of sectors in the polar mask layer as four. 
\\
\textbf{Effect of RSAS.}
The proposed RSAS is employed to extract deep features along the radius and circumference, corresponding to modeling translation and rotation motion information in the context of blurry images.
As shown in Table~\ref{tab:abl-module}, we use the Swin Transformer as the base architecture for comparison.
In the last row of Table~\ref{tab:abl-module}, DRE and RSAS boost the performance of 1.86 dB compared with the baseline model.
Combined with RSAS, the RST performs a significant performance of the sharp image restoration, demonstrating the effectiveness of the module.
The main reason is that RSAS considers the rotation and translation motion information together, which more adequately explores the context of blur pattern.
The visual results are shown in Figure~\ref{pic:abl_fft}.
We conduct the experiment directly using the Swin Transformer~\cite{liu2021swin} on our RST architecture, which performs a worse image deblurring result, as shown in Figure~\ref{pic:abl_swin}.
\\
 \textbf{Model Efficiency.}
The computational complexity comparison is shown in Table~\ref{table4time}, RST achieves the lowest parameters and the second lowest FLOPs compared with 8 state-of-the-art methods~\cite{eccv2022_Stripformer,kong2023efficient,fang2023self}.
We have the less FLOPs than the latest SOTA method FFTformer~\cite{kong2023efficient} and UFPNet~\cite{fang2023self}.
To other previous works, RST has less or comparable complexity with superior performance.
The above results well prove the effect and efficiency of RST. 
\begin{table}[!t]\footnotesize
\centering
\caption{The computational complexity comparison with the input size of $256 \times 256$ in terms of Params (M) and FLOPs (G).
}
\footnotesize
\scalebox{0.750}{
\setlength{\tabcolsep}{0.8mm}
{
\begin{tabular}{cccccccccccccccccccc}

\toprule[0.8pt]
\multicolumn{2}{l|}{} 
&\multicolumn{2}{c}{ DeblurGAN-v2}
& \multicolumn{2}{c}{MIMO} 
& \multicolumn{2}{c}{MPRNet} 
& \multicolumn{2}{c}{Restormer}
& \multicolumn{2}{c}{Uformer}
& \multicolumn{2}{c}{Stripformer}
& \multicolumn{2}{c}{FFTformer}
& \multicolumn{2}{c|}{UFPNet}
& \multicolumn{2}{c}{RST}
\\ 

\multicolumn{2}{l|}{\textbf{Method}} 
&\multicolumn{2}{c}{\cite{kupyn2019deblurgan}}
& \multicolumn{2}{c}{\cite{cvpr2021cho}} 
& \multicolumn{2}{c}{\cite{zamir2021multi}} 
& \multicolumn{2}{c}{\cite{zamir2022restormer}}
& \multicolumn{2}{c}{\cite{wang2022uformer}}
& \multicolumn{2}{c}{\cite{eccv2022_Stripformer}}
& \multicolumn{2}{c}{\cite{kong2023efficient}}
& \multicolumn{2}{c|}{\cite{fang2023self}}
& \multicolumn{2}{c}{(Ours)}
\\ 

\midrule[0.8pt]
\multicolumn{2}{l|}{FLOPs (G)}           
 & \multicolumn{2}{c}{411.34} 
& \multicolumn{2}{c}{154.58} & \multicolumn{2}{c}{777.68}
& \multicolumn{2}{c}{141.24} & \multicolumn{2}{c}{\textbf{89.46}}
& \multicolumn{2}{c}{177.65} & \multicolumn{2}{c}{131.75}
& \multicolumn{2}{c|}{243.78} & \multicolumn{2}{c}{\underline{112.48}}\\
\multicolumn{2}{l|}{Params (M)}           
 & \multicolumn{2}{c}{60.9} 
& \multicolumn{2}{c}{\underline{16.1}} & \multicolumn{2}{c}{20.1}
& \multicolumn{2}{c}{26.1} & \multicolumn{2}{c}{50.9}
& \multicolumn{2}{c}{19.7} & \multicolumn{2}{c}{16.6}
& \multicolumn{2}{c|}{80.3} & \multicolumn{2}{c}{\textbf{14.3}}\\
\bottomrule[0.8pt]
\end{tabular}
}
}
\label{table4time}
\end{table}
\section{Conclusion}
In this paper, we propose an efficient and effective transformer-based model with the polar coordinate system for the image deblurring task, called RST.
To obtain the relationship between tokens in the slanting directions and magnitudes, RST utilizes two parts: DRE and RSAS to demand not only less computation than previous works but also excellent performance.
The experimental results show that our model achieves state-of-art performance on five synthesis and real-world datasets(\eg, GoPro, HIDE, RealBlur, REDS, RSBlur). \\
\textbf{Limitation.} Our method still has some shortcomings for future improvement. 
Although we designed the radial strip window to reduce the computational cost, which also leads the insufficient cross-window interactions.
For the heavy blurry in overly complex real-world scenarios, our proposed method still has limited capacity, as shown in the failure cases in supplemental materials.
Creating a dataset that can cover the various real-world might be a solution for improvements

\bibliographystyle{splncs04}
\bibliography{egbib}
\end{document}